%% file: main.tex
\documentclass[letterpaper, 10 pt, conference]{ieeeconf}
\IEEEoverridecommandlockouts    
\usepackage{subfigure}
\usepackage[table]{xcolor}
\usepackage{graphicx}
\usepackage{textcomp}
\usepackage{tikz}
\usepackage{amsmath,epsfig}
\usepackage{verbatim}
\usepackage{lipsum,multicol}
\usepackage{dblfloatfix}
\usepackage{url}
\usepackage{cite}
\usepackage{tabularx}
\usepackage{tabulary}
\usepackage{amsfonts}
\usepackage{amssymb}
\usepackage{gensymb}
\usepackage{float}
\usepackage{balance}
\usepackage{booktabs}

\usepackage{multirow}
\usepackage{verbatimbox}

\newcommand{\steve}[1]{{#1}}

\title{\LARGE \bf
A Normal Distribution Transform-Based Radar Odometry Designed For Scanning and Automotive Radars}

\author{Pou-Chun Kung, Chieh-Chih Wang and Wen-Chieh Lin

\thanks{Pou-Chun Kung is with the Graduate Degree Program of Robotics, National Chiao Tung University, Hsinchu, Taiwan. 
        {E-mail: pouchunkung@gmail.com}}%
\thanks{Chieh-Chih Wang is with the Department of Electrical and Computer Engineering, National Chiao Tung University, and with the Mechanical and Mechatronics Systems Research Laboratories, Industrial Technology Research Institute, Hsinchu, Taiwan.    
        {E-mail: bobwang@ieee.org}}%
\thanks{Wen-Chieh Lin is with the Department of Computer Science, National Chiao Tung University, Hsinchu, Taiwan. 
        {E-mail: wclin@cs.nctu.edu.tw}}%
}

\date{February 2020}

\begin{document}

\maketitle

\thispagestyle{empty}
\pagestyle{empty}

\begin{abstract}
Existing radar sensors can be classified into automotive and scanning radars. While most radar odometry (RO) methods are only designed for a specific type of radar, our RO method adapts to both scanning and automotive radars. Our RO is simple yet effective, where the pipeline consists of thresholding, probabilistic submap building, and an NDT-based radar scan matching. The proposed RO has been tested on two public radar datasets: the Oxford Radar RobotCar dataset and the nuScenes dataset, which provide scanning and automotive radar data respectively. 
The results show that our approach surpasses state-of-the-art RO using either automotive or scanning radar by reducing translational error by 51\% and 30\%, respectively, and rotational error by 17\% and 29\%, respectively.
Besides, we show that our RO achieves centimeter-level accuracy as lidar odometry, and automotive and scanning RO have similar accuracy.
\end{abstract}

\begin{keywords}
Radar, Odometry, Scan Matching, Autonomous Driving.
\end{keywords}

\section{Introduction}

\label{sec:introduction}

\begin{figure}[t]
    \centering
        \includegraphics[width = 1.3in]{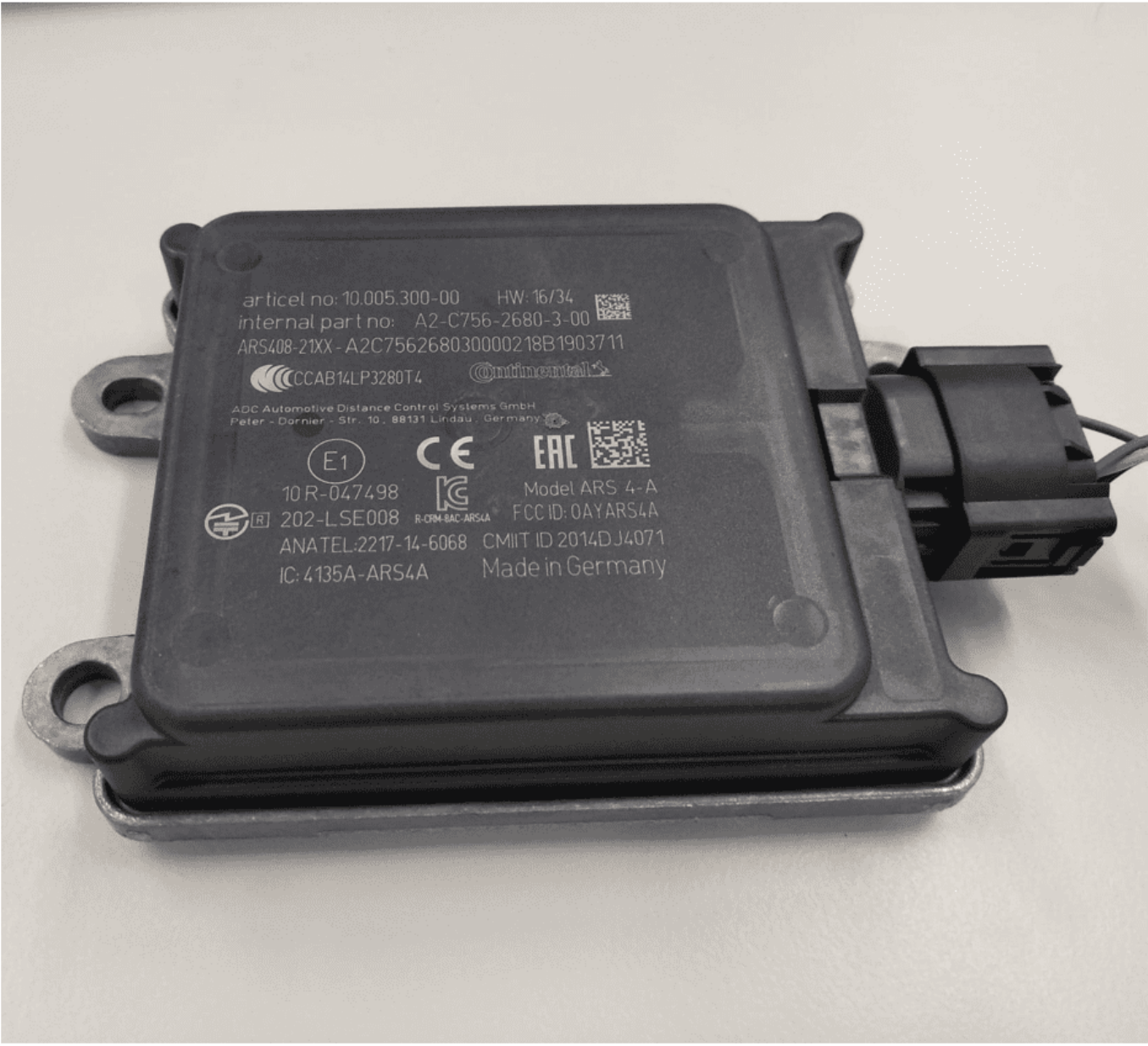}
        \includegraphics[width = 1.2in]{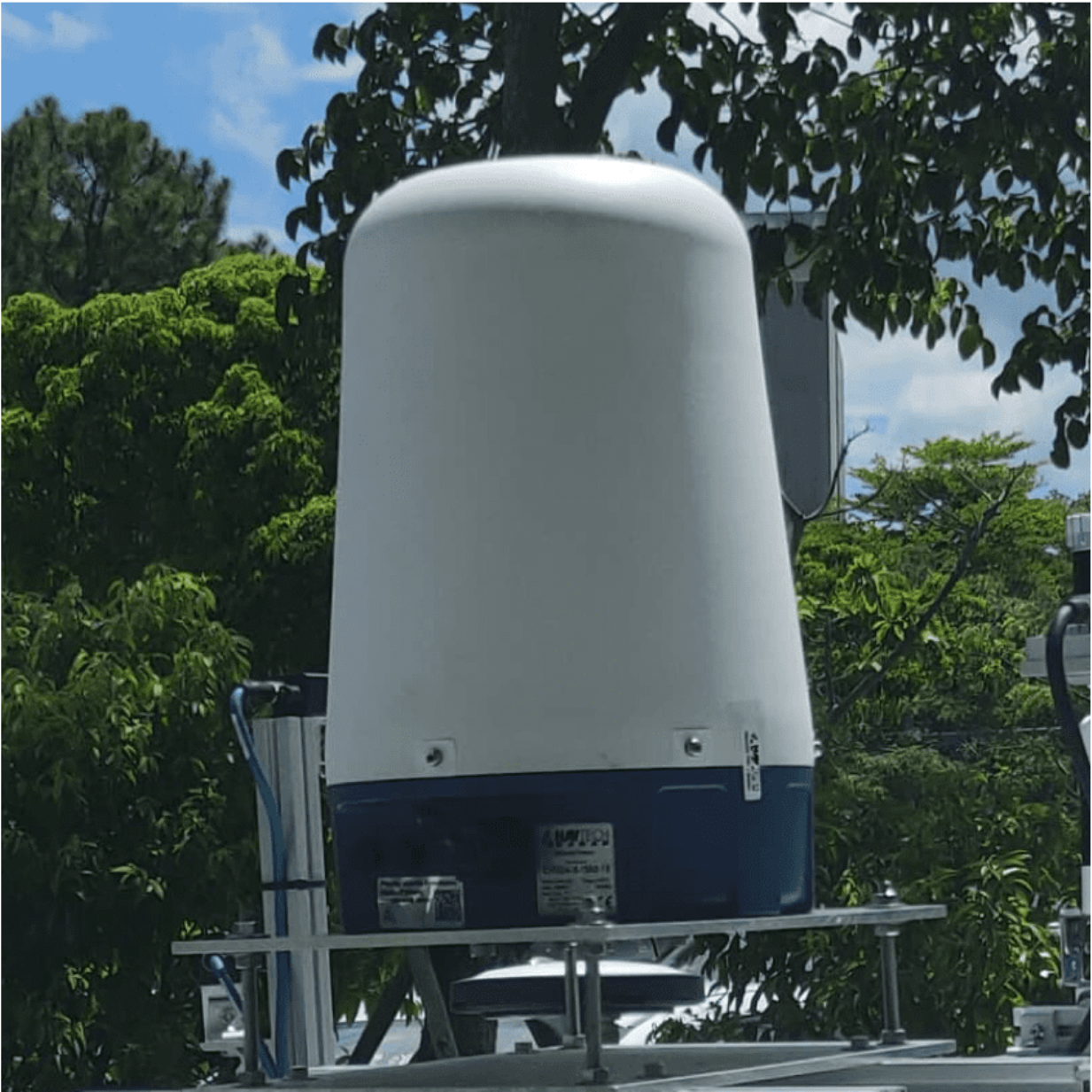}
        \vspace{-0.2cm}
        \caption{Pictures of an automotive radar (left) and a scanning radar (right).}
        \label{fig:radars}
    \end{figure}

\begin{figure}[t]
    \centering
        \includegraphics[width = 1.25in]{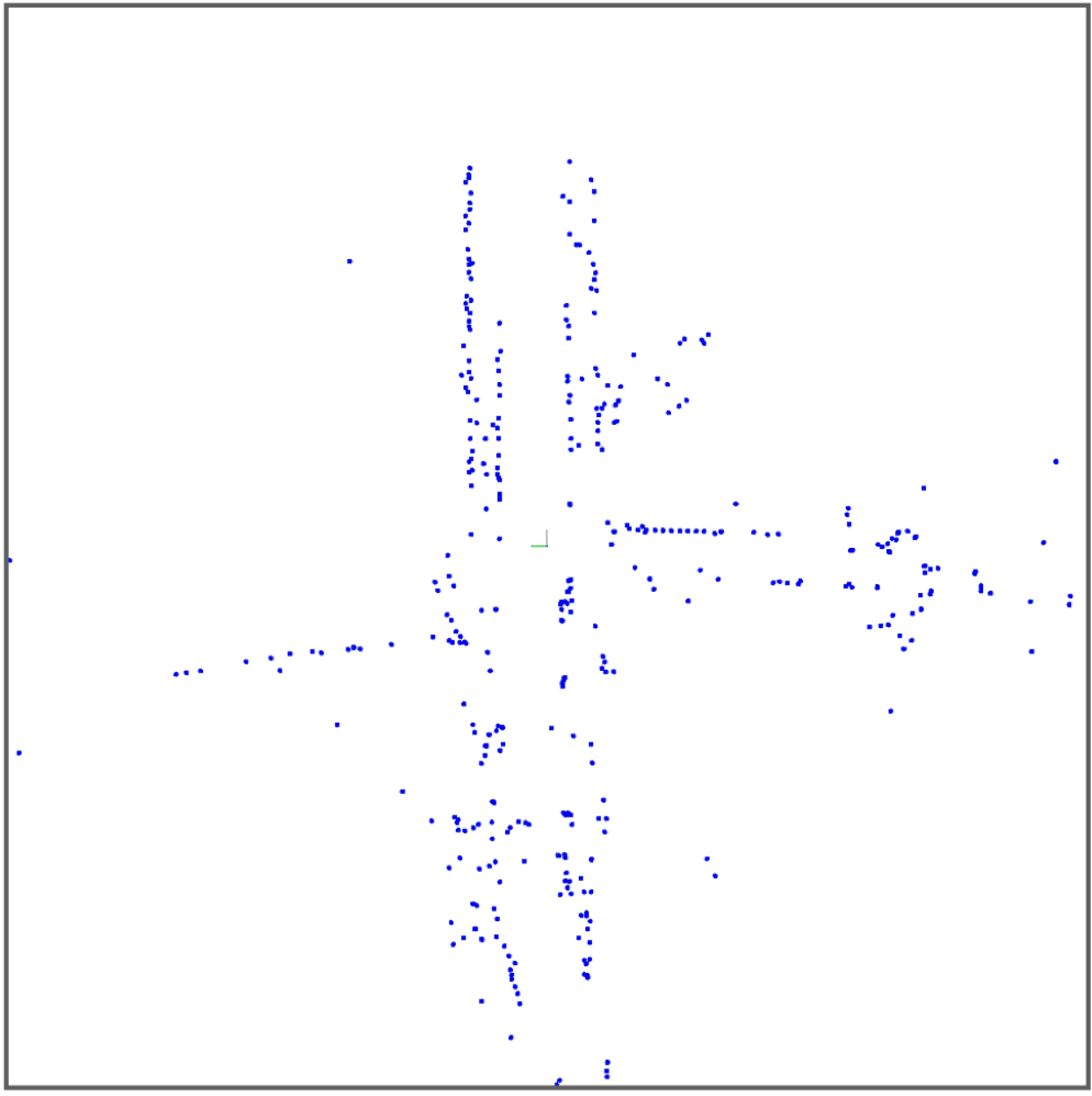}
        \includegraphics[width = 1.25in]{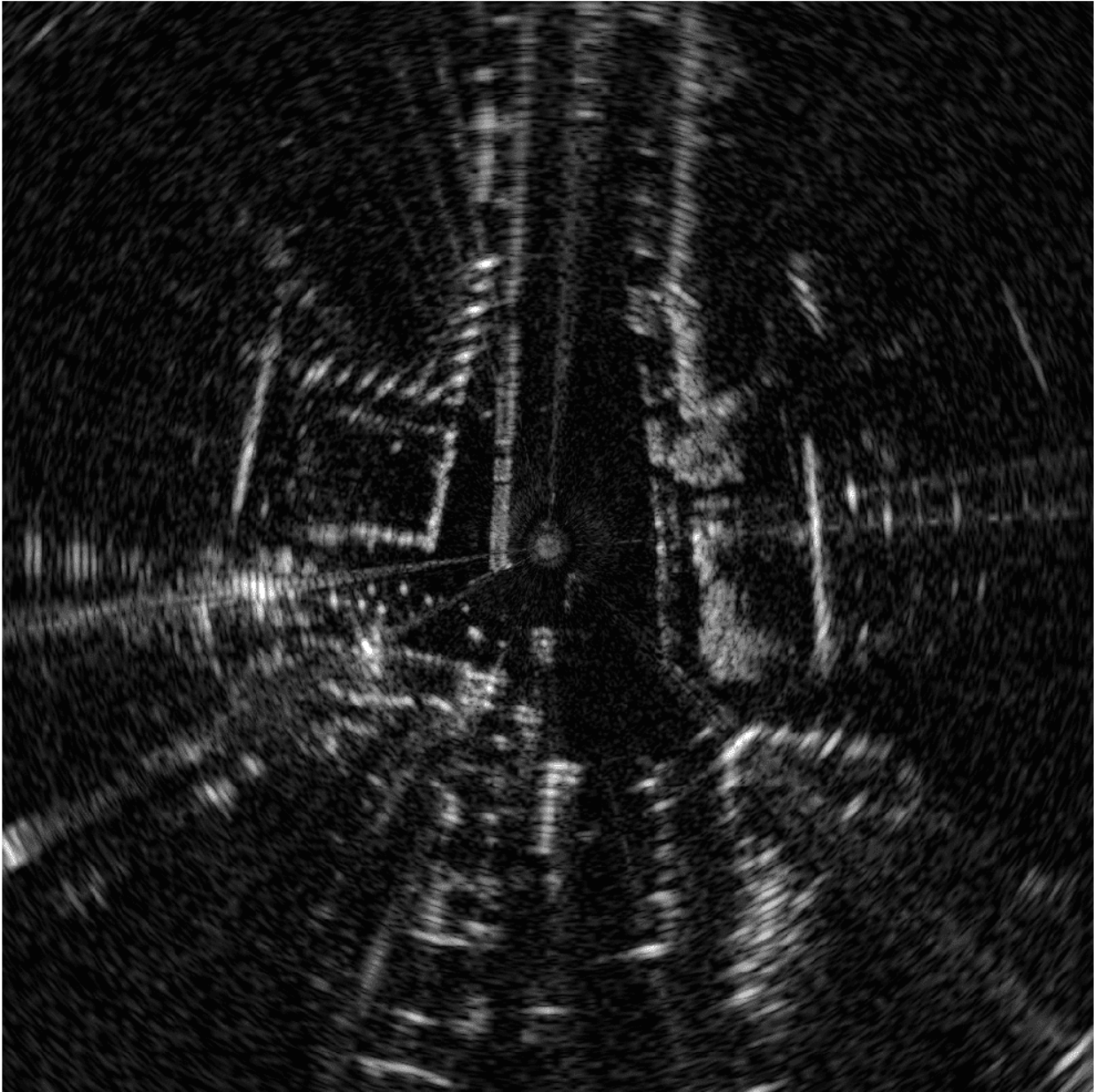}
        \vspace{-0.2cm}
        \caption{Pictures of a 360-degree radar scan provided by five automotive radars with different facing directions (left) and a radar scan by a scanning radar (right).}
        \vspace{-0.2cm}
        \label{fig:radar scans}
    \end{figure}

Radar sensors could be categorized into automotive radars and scanning radars. With Doppler information, automotive radars offer radial velocity measurements. 
However, measurements from automotive radars are with relatively low accuracy and are sparse since only highly confident measurements are output.
In contrast, scanning radars provide raw power-range images with relatively high angular and range resolution. However, radar images from scanning radars include noise and do not provide velocity information. 
Fig.~\ref{fig:radars} shows an automotive radar from Continental and a scanning radar from NavTech, whereas sample measurements from these radars are shown in Fig.~\ref{fig:radar scans}. The specifications of the automotive and scanning radar are listed in Table \ref{tab:radar table} as a reference to their radar performance characteristics.  


In this paper, we present the first radar odometry (RO) method can operate on both automotive and scanning radar. The automotive radar measurements are sparse but more likely to be the real-world object, whereas the scanning radar scans are dense but contain more noise. 
The proposed RO aims to reconstruct dense information from automotive radar scans and remove noise from scanning radar scans by the proposed probabilistic radar submap and thresholding module, respectively. Finally, RO is achieved by consecutive radar scan matching using the proposed NDT-based method.

Our RO is simple but effective. Table \ref{tab:Experiment} shows that the proposed RO surpasses state-of-the-art (SOTA) automotive and scanning RO.
While SOTA automotive RO \cite{holder2019real} proposed submap matching to solve the sparseness issue and significantly improve the performance of automotive RO, the proposed RO outperforms \cite{holder2019real} by extending the submap to probabilistic submap, which takes uncertainties into account. 
On the other hand, our method 
surpasses SOTA learning-based scanning RO \cite{barnes2019masking} \steve{in that} the proposed NDT-based matching \steve{can better handle outliers than the correlation-based matching utilized in \cite{barnes2019masking}, even though we simply remove radar noise by thresholding (rather than a trained masking network used in \cite{barnes2019masking}).} 
The proposed RO is also easy-to-use because it does not require training.

\input{tables/Radar_Table}

We further show a comparison between the proposed RO and the lidar odometry achieved by consecutive ICP scan matching. 
For a fair comparison, all sensors are downsampled to approximately 4Hz, and the frame-by-frame errors are calculated. 
The results of the RO and the lidar odometry on two public datasets are shown at the bottom of the Table \ref{tab:Experiment}. 
Both the automotive and scanning RO achieve centimeter-level accuracy as lidar odometry does. And the automotive RO shows competitive accuracy with scanning RO, although the automotive radar is with relatively low accuracy and resolution.



\section{Related Work}
\label{sec:Related Work}

\subsection{Automotive Radar Odometry}

Early automotive RO proposed by Kellner et al. \cite{kellner2013instantaneous,kellner2014instantaneous} used radial velocity measurements to estimate ego-velocity and removed outliers by RANSAC \cite{fischler1981random} according to the radial velocity of each radar point.
Later methods applied radar point cloud matching, e.g., matching radar point clouds by Binary Annular Statistics Descriptor (BASD) visual descriptor \cite{schuster2016landmark}, ICP matching \cite{ward2016vehicle}, NDT-based matching \cite{rapp2015fast,rapp2017probabilistic}, and occupancy map matching \cite{rapp2015clustering}. Rapp et al. \cite{rapp2015fast,rapp2017probabilistic} applied the joint-Doppler clustering-based NDT, which weights each radar point by the similarity between measured velocity and the estimated velocity to reduce the outlier effect. However, the methods based on the point cloud matching method could fail when the radar point cloud is too sparse. Holder et al. \cite{holder2019real} proposed a radar submap to handle the sparseness issue and achieved RO by consecutive radar submap matching by ICP, where the submap is the combination of several radar scans to construct more information in a single radar submap. However, this method does not consider the uncertainty of ego-motion used to build submap and radar measurements and requires an external sensor instead of using radar only.

\subsection{Scanning Radar Odometry}

For the scanning radar offering power-range images, feature-based scanning RO methods were proposed. Earlier researchers leveraged constant false alarm rate (CFAR) \cite{rohling1983radar} to extract features and associate features by minimizing a cost function defined by scanning radar's distortion model \cite{vivet2013localization}. Vision-inspired work \cite{callmer2011radar,aldera2019could,aldera2019fast,hong2020radarslam} utilized image feature extractors and descriptors such as SIFT, FAST and SUFT on radar images. Cen et al. \cite{ro_cen1,ro_cen2} proposed a new feature extraction approach and descriptor for RO and used a shape similarity metric to remove outliers. Other work \cite{burnett2021we} consider motion distortion and Doppler effect in RO.
Recently, learning-based methods were applied to extract features from a radar image \cite{barnes2020under,suaftescu2020kidnapped,de2020kradar}. 
\steve{Although the aforementioned feature-based methods attempted to extract features that can represent real-world objects in noisy radar images, they also drops the dense information provided by scanning radar.}

\input{tables/Experiment}

The correlation-based approaches have also been used widely in scanning RO. Earlier studies utilized Fourier-Mellin transformation (FMT), maximizing the phase correlation between two radar images in Fourier domain \cite{checchin2010radar}. A sensing-distortion analysis was applied to increase the accuracy of FMT \cite{vivet2012radar,vivet2013mobile}. To improve FMT's robustness, \steve{PhaRao \cite{park2020pharao}  
added a keyframe selection process to drop frames that are seriously affected by outliers. Dropping frames, however, could lose information and not the best way to handle outliers.} The SOTA scanning RO proposed by Barnes et al. \cite{barnes2019masking} accomplished odometry by searching the best correlation between consecutive radar images with a mask provided by a trained network to remove outliers, such as moving objects and radar noise. The masking network shows significant improvement in the accuracy of RO by reducing errors over 25\% in translation and 11\% in rotation error from the former study \cite{ro_cen2}. A main limitation of Barnes et al.'s approach is that correlative scan matching could be affected by outliers if the masking network did not remove all outliers.

\section{Method}
\label{sec:Method}

The proposed RO method consists of: (i) a thresholding module, (ii) a probabilistic submap module, (iii) a NDT module, and (iv) a NDT matching module. Each module is described in the following subsections. In the end, we discuss the influence of each module. 
	
	
\subsection{Thresholding}

\begin{figure}[t]
	\vspace{5pt} 
	\includegraphics[width = 1.5in]{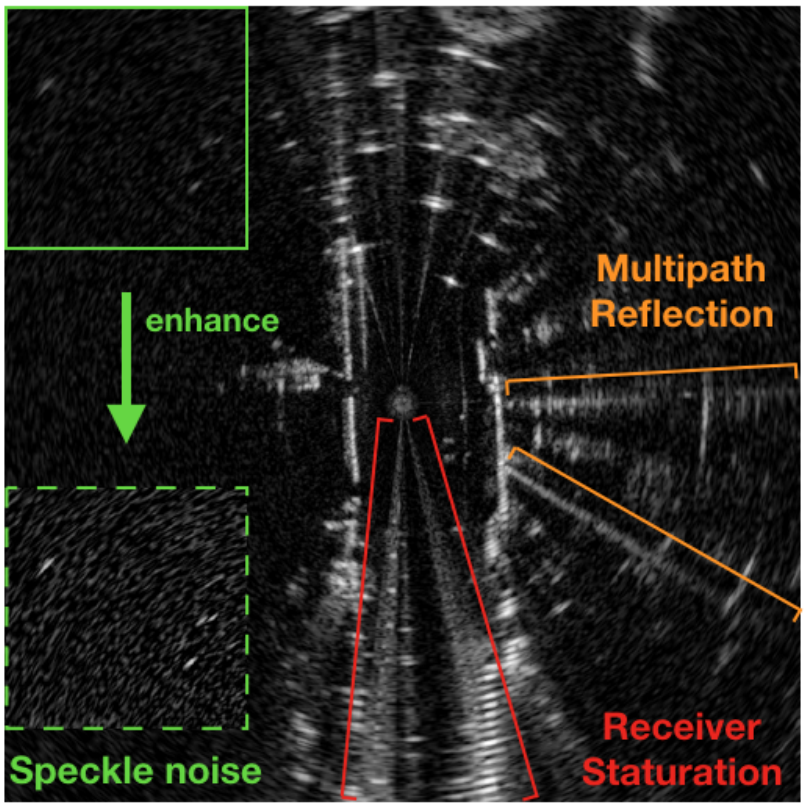}
	\centering
	\vspace{-0.3cm}
	\caption{\steve{Three typical radar noise} in a radar image: speckle noise, multipath reflection, and receiver saturation.}
	\label{fig:radar noise}
	\end{figure}

\begin{figure}[t]
	\includegraphics[width = 3.4in]{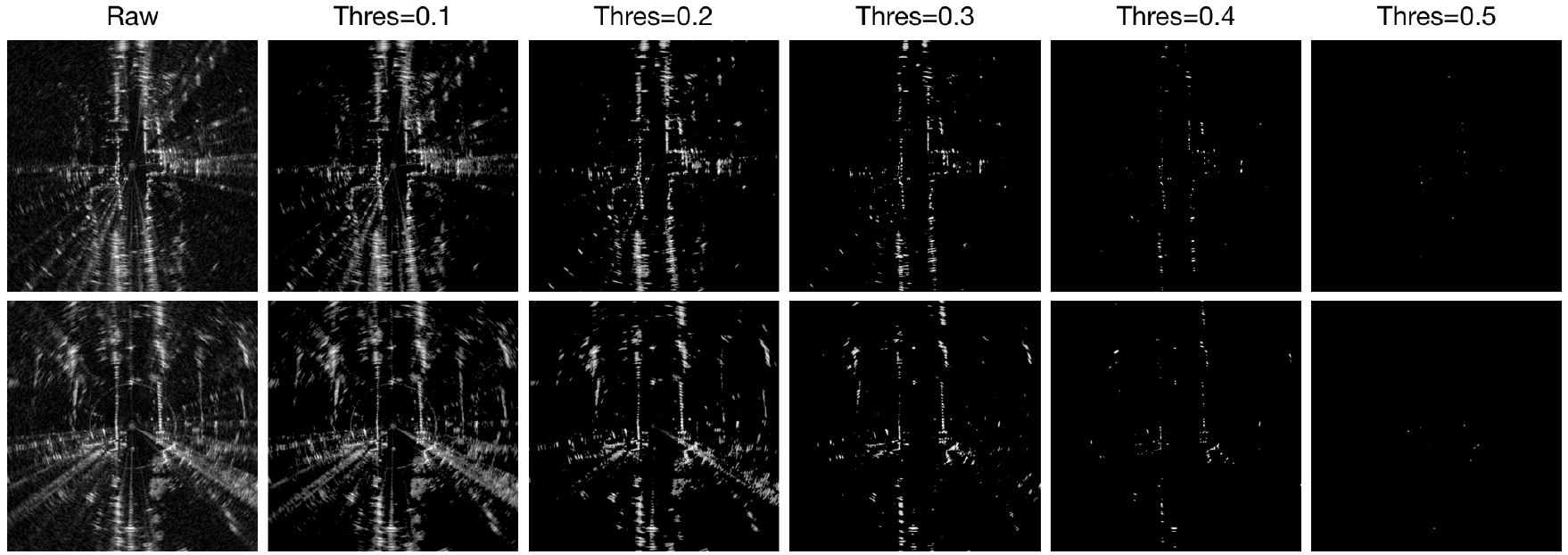}
	\centering
	\caption{\steve{Radar images processed with different thresholds. In Oxford dataset, most of the noise can be removed by setting a threshold value higher than 0.3. However, the radar images become sparser as the threshold increases.}}
	\label{fig:thres}
	\end{figure}

Radar noise is a well-known artifact in radar measurements. Three typical types of radar noise are shown in Fig.~\ref{fig:radar noise}. The system on chip (SoC) in automotive radar is designed to extract radar points and remove ghost detection caused by radar noise. However, the radar image provided by scanning radar is a raw power-range image including radar noise, \steve{so this issue needs to be particularly solved in scanning RO. 
The} masking method in \cite{barnes2019masking} outperforms most of the existing scanning RO \cite{ro_cen2}. Inspired by this prior art, we propose a straightforward thresholding method to reduce radar noise because we found that most of the noise has relatively low returned power. Fig.~\ref{fig:thres} illustrates that thresholding can successfully reduce radar noise. In this paper, a fixed threshold value is set for all radar scans. We tune the thresholding parameter using one sequence of Oxford dataset and test on the all sequences.

\subsection{Probabilistic Radar Submap}

\begin{figure}[t]
	\vspace{2pt} 
	\includegraphics[width = 3.1in]{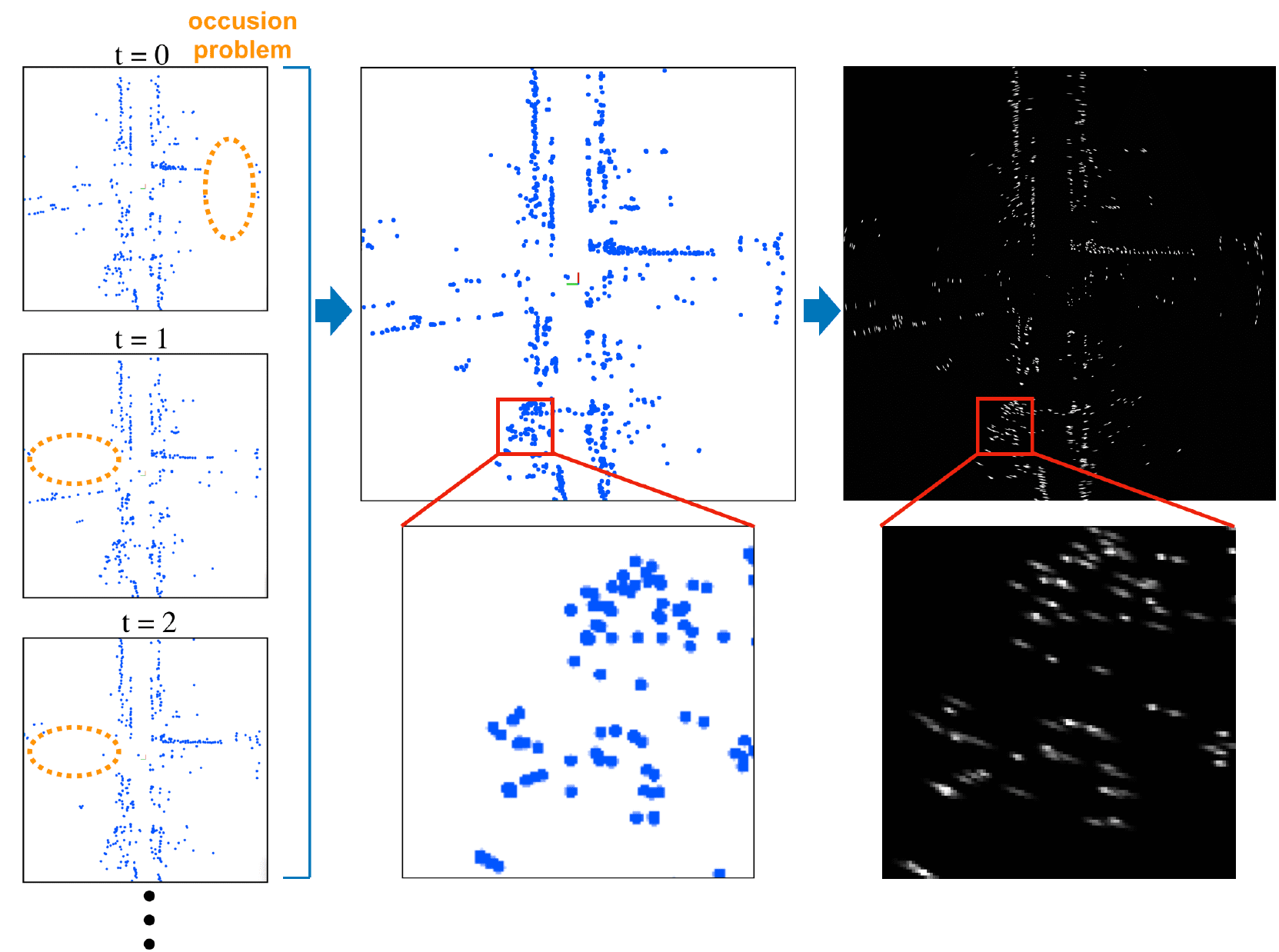}
	\centering
    \caption{\steve{An example illustrates our probabilistic radar submap (right image) built from three automotive radar scans (left images). Our result can better represent the distribution of radar measurements than the radar submap \cite{holder2019real} (middle image). Brighter intensity represents a higher probability. 
    } 
    }
	\label{fig:submap}
	\end{figure}

Inspired by radar submap matching in \cite{holder2019real}, we proposed a probabilistic radar submap that not only combines 
\steve{multiple} radar scans into the radar submap but also takes the uncertainty of measurements and the ego-motion for building the submap into account. To construct the probabilistic radar submap \steve{using automotive radars}, the ego-velocity estimation \cite{kellner2014instantaneous} calculated by Doppler velocity measurements is deployed. \steve{For scanning radars, the ego-velocity cannot be estimated using \cite{kellner2014instantaneous}. In this case, IMU integration is a possible solution to obtain the ego-velocity, which then can be used to build a probabilistic radar submap for scanning RO. To evaluate the radar-only odometry performance, however, the probabilistic radar submap was not applied to scanning RO in our experiment. Fig.~\ref{fig:submap} shows a probabilistic radar submap presented by a sparse Gaussian Mixture Model.} 

\begin{figure}[t]
        \centering
        \subfigure[]{
        \includegraphics[width = 1.3in]{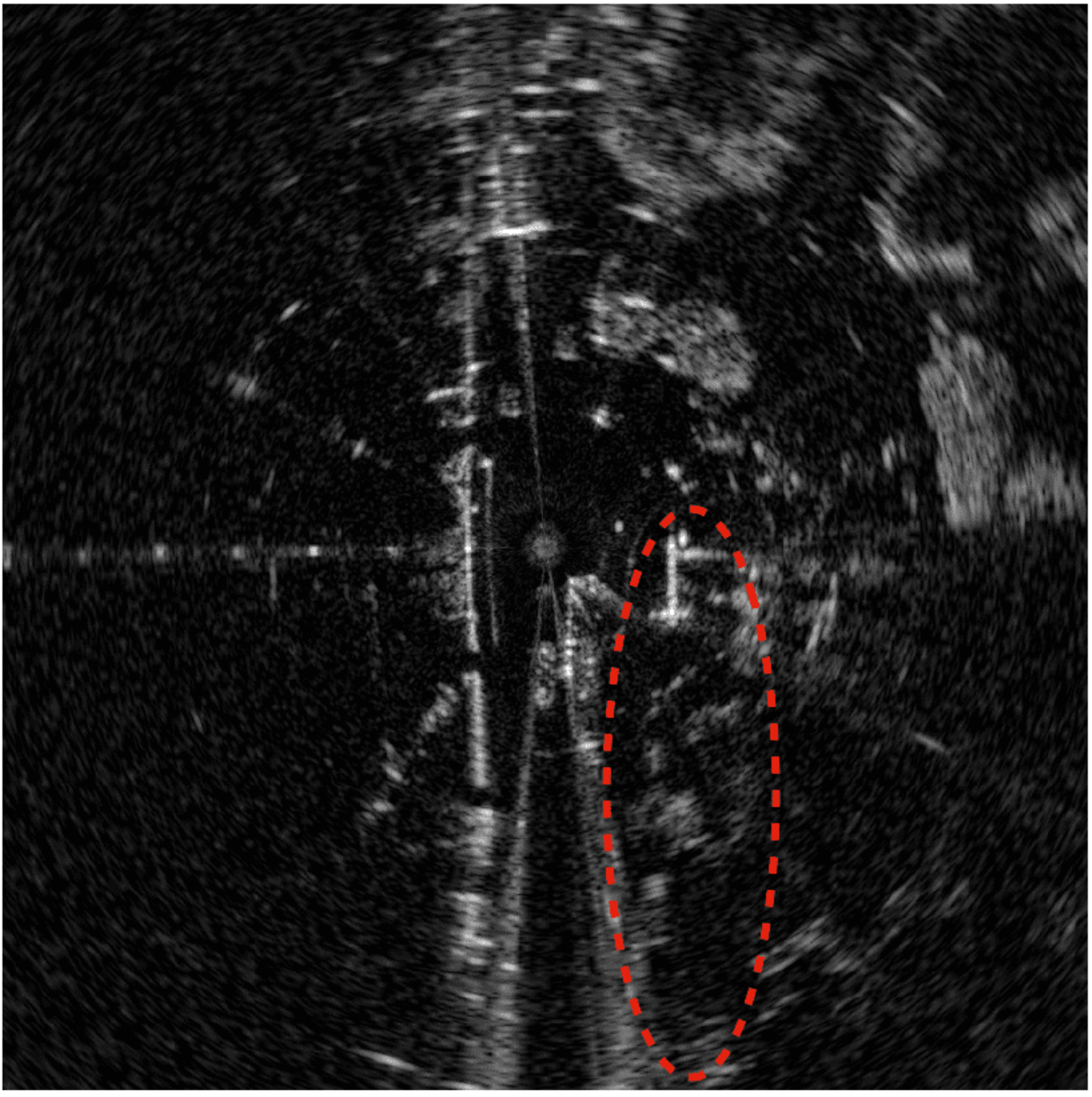}
        \label{fig:occlusion_scanning_a}
        }
        \subfigure[]{
        \includegraphics[width = 1.3in]{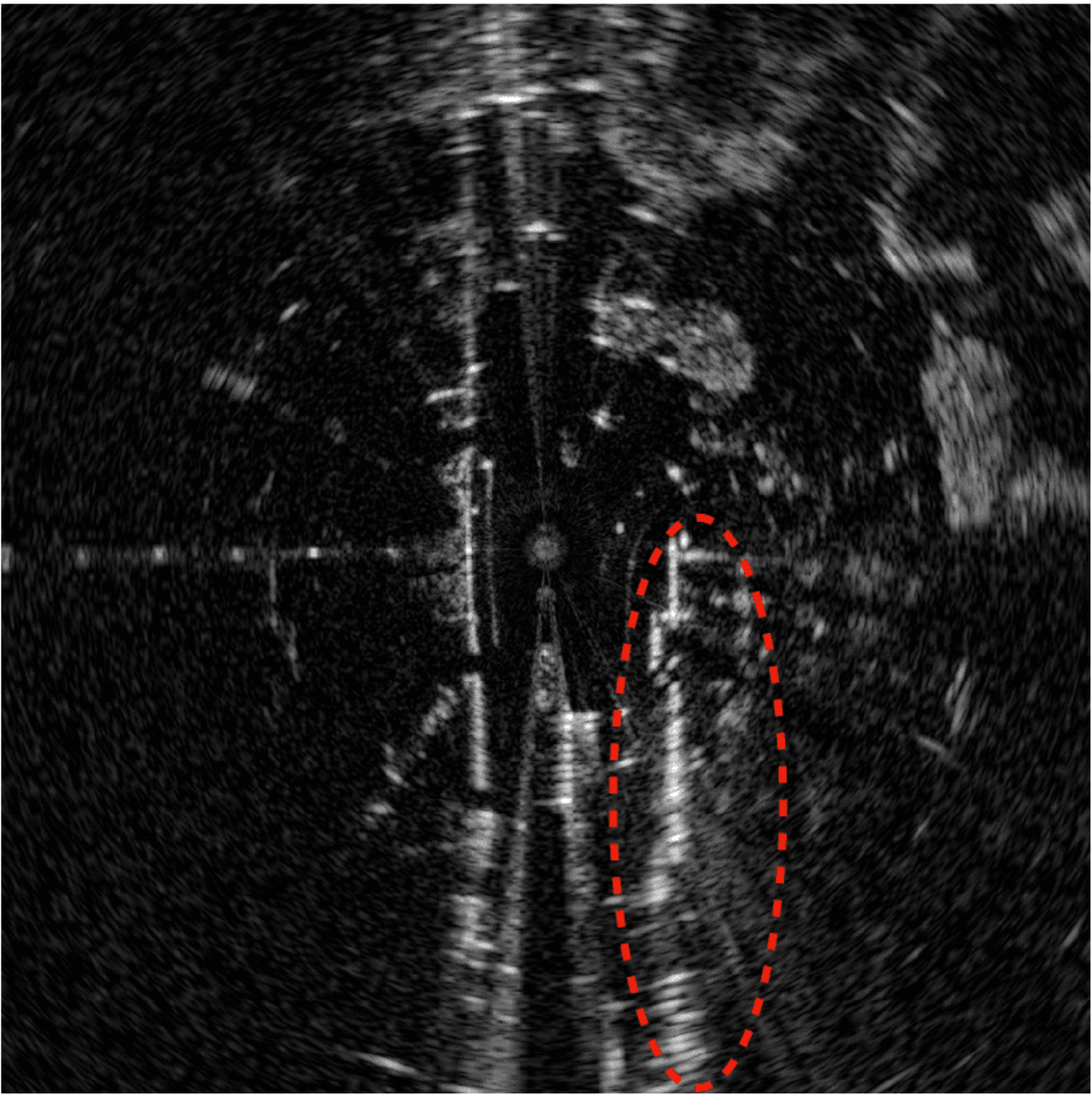}
        \label{fig:occlusion_scanning_b}
        }
        \caption{An example of occlusion occurs in scanning radars. (a) Radar scan with a building (red circled) occluded by a bus. (b) Radar scan of the same building after the bus left.}
        \label{fig:occlusion_scanning} 
    \end{figure}

Although radar is not an optical sensor, occlusion still occurs occasionally when the radar wave is blocked by those objects having low penetrability or high reflectivity, as shown in Fig.~\ref{fig:submap} and Fig.~\ref{fig:occlusion_scanning}. The probabilistic submap can handle not only the sparseness problem of automotive radars but also the occlusion problem of radars. 

\subsection{Normal Distribution Transform}

We propose a weighted probabilistic NDT. It converts a radar scan into a set of the local normal distribution functions according to uncertainties and the returned power of radar points. Each normal distribution function describes the shape of a part of the radar scan. \steve{The first step is to regularly subdivide the space occupied by the radar scan into grids of size $g \times g$ meters.} 
Then the local normal distribution for each gird is computed as the probabilistic NDT shown in \cite{hong2017probabilistic}, and each radar point is weighted by the shifted returned power $p_i$,

\begin{equation}
w_i = p_i - s
\end{equation}
where $s$ is a parameter used to shift the weights of radar points based on their returned power. If the weight after shifting is negative, then the weight is set to zero.

\begin{figure}[t]
    \vspace{5pt} 
        \centering
        \subfigure[]{
        \includegraphics[width = 1.1in]{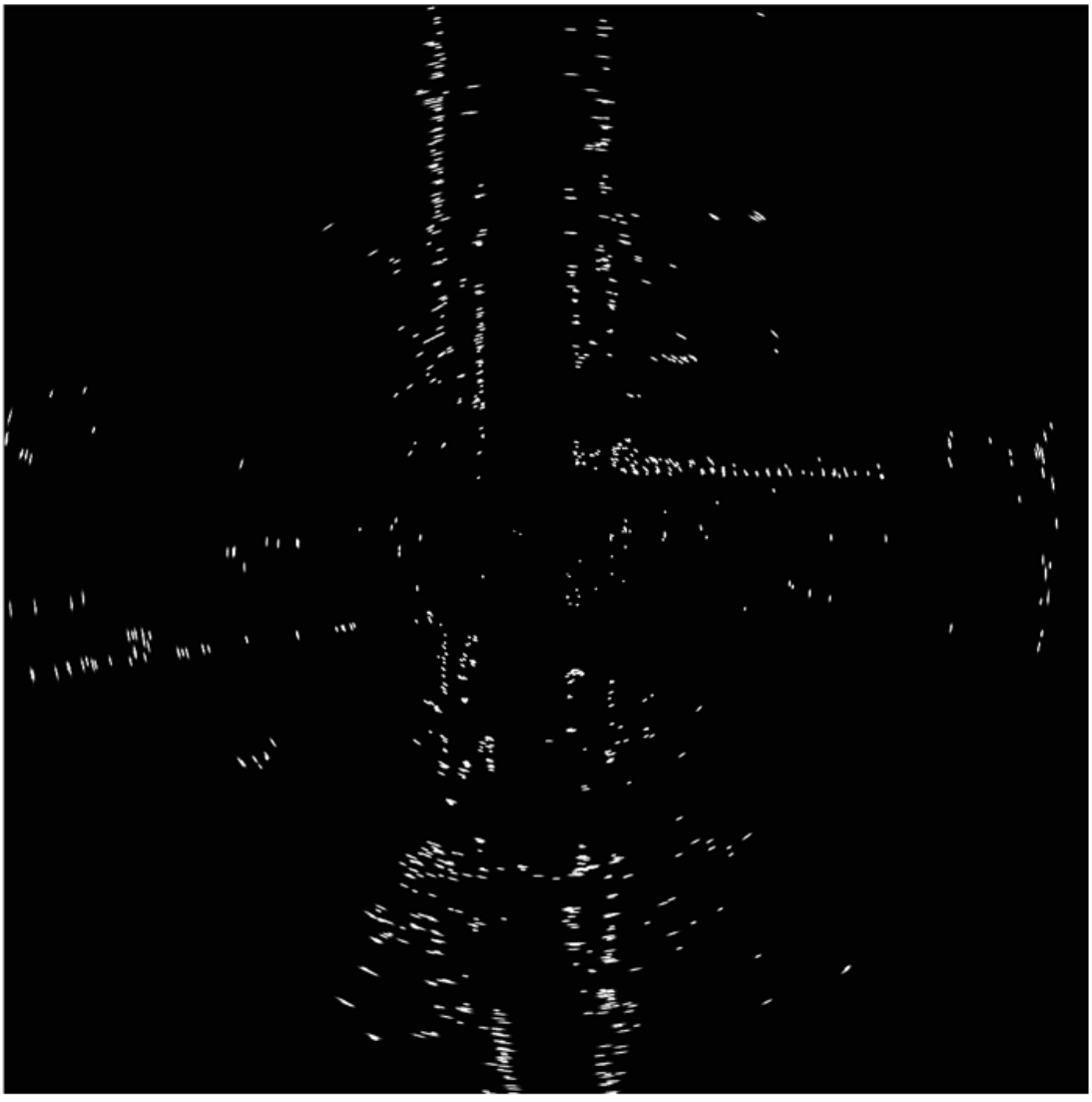}
        \label{fig:sparse_nd_map_a}
        }\hspace{-0.14in}
        \subfigure[]{
        \includegraphics[width = 1.1in]{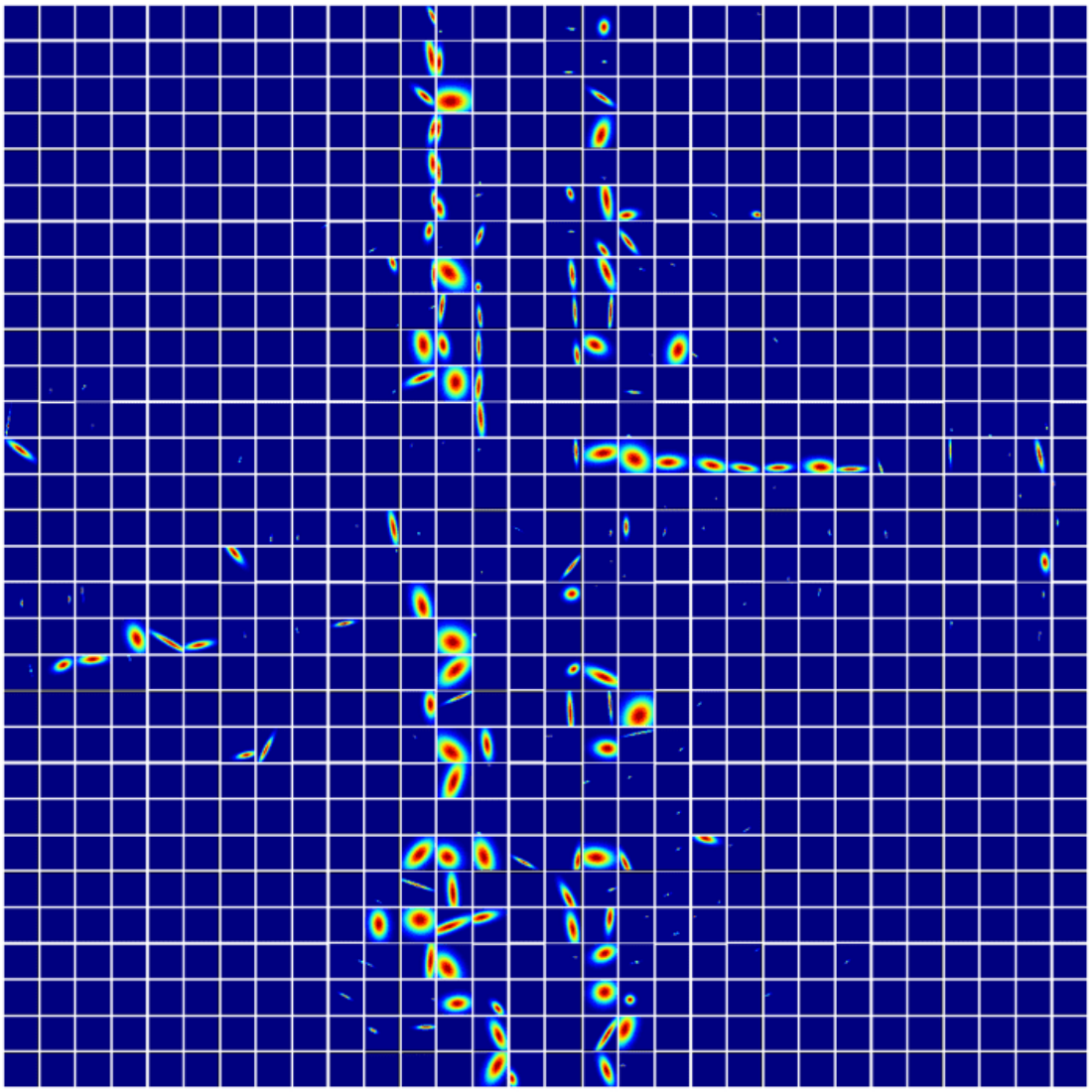}
        \label{fig:sparse_nd_map_b}
        }\hspace{-0.14in}
        \subfigure[]{
        \includegraphics[width = 1.1in]{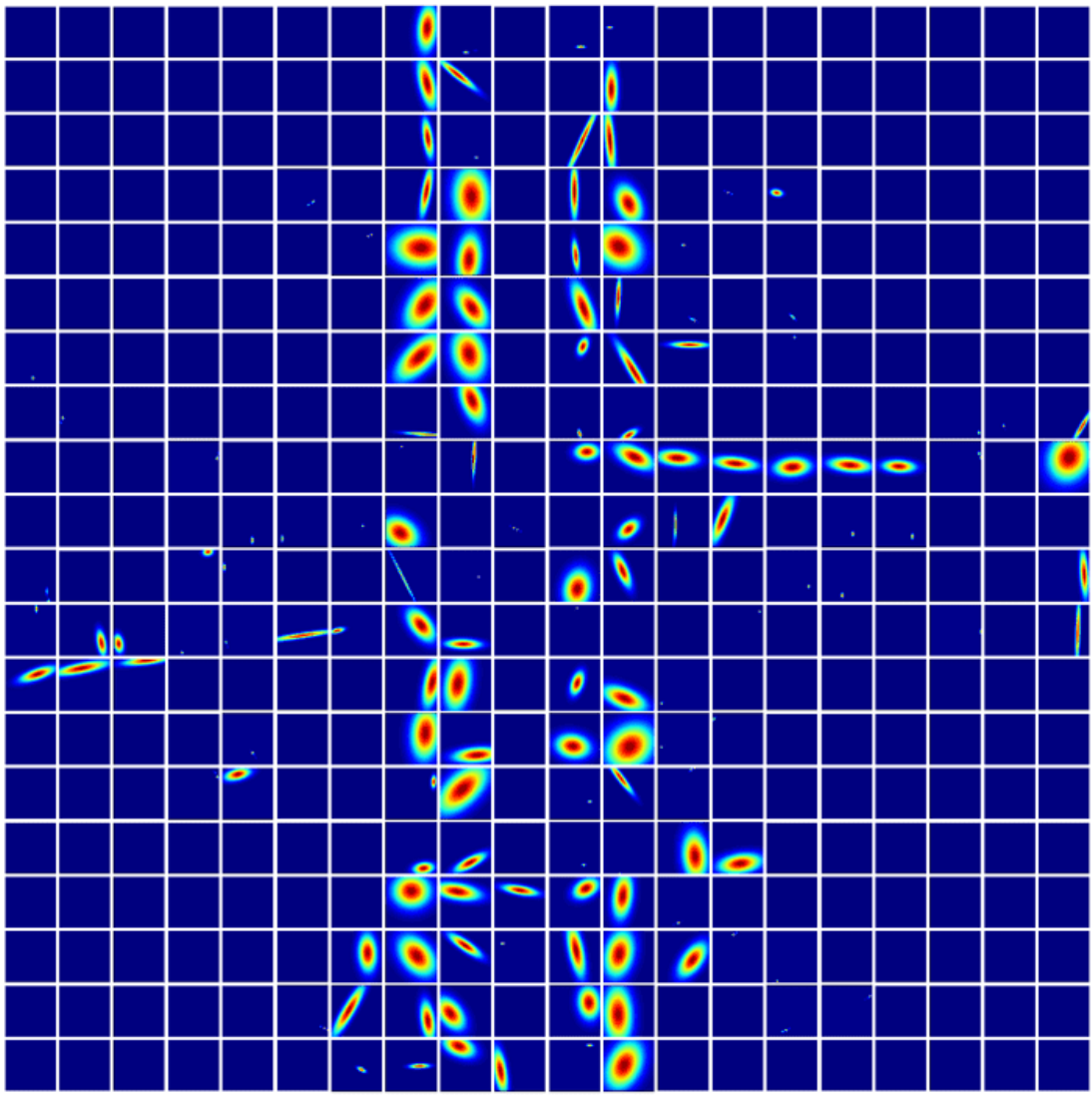}
        \label{fig:sparse_nd_map_c}
        }\hspace{-0.14in}
        \caption{Illustration of the normal distribution map created from a automotive radar scan. (a) The probabilistic radar submap. (b),(c) The normal distribution map with the grid size of 5 m and 7 m for visualization.}
        \label{fig:ND Map sparse} 
    \end{figure}

\begin{figure}[t]
        \centering
        \subfigure[]{
        \includegraphics[width = 1.1in]{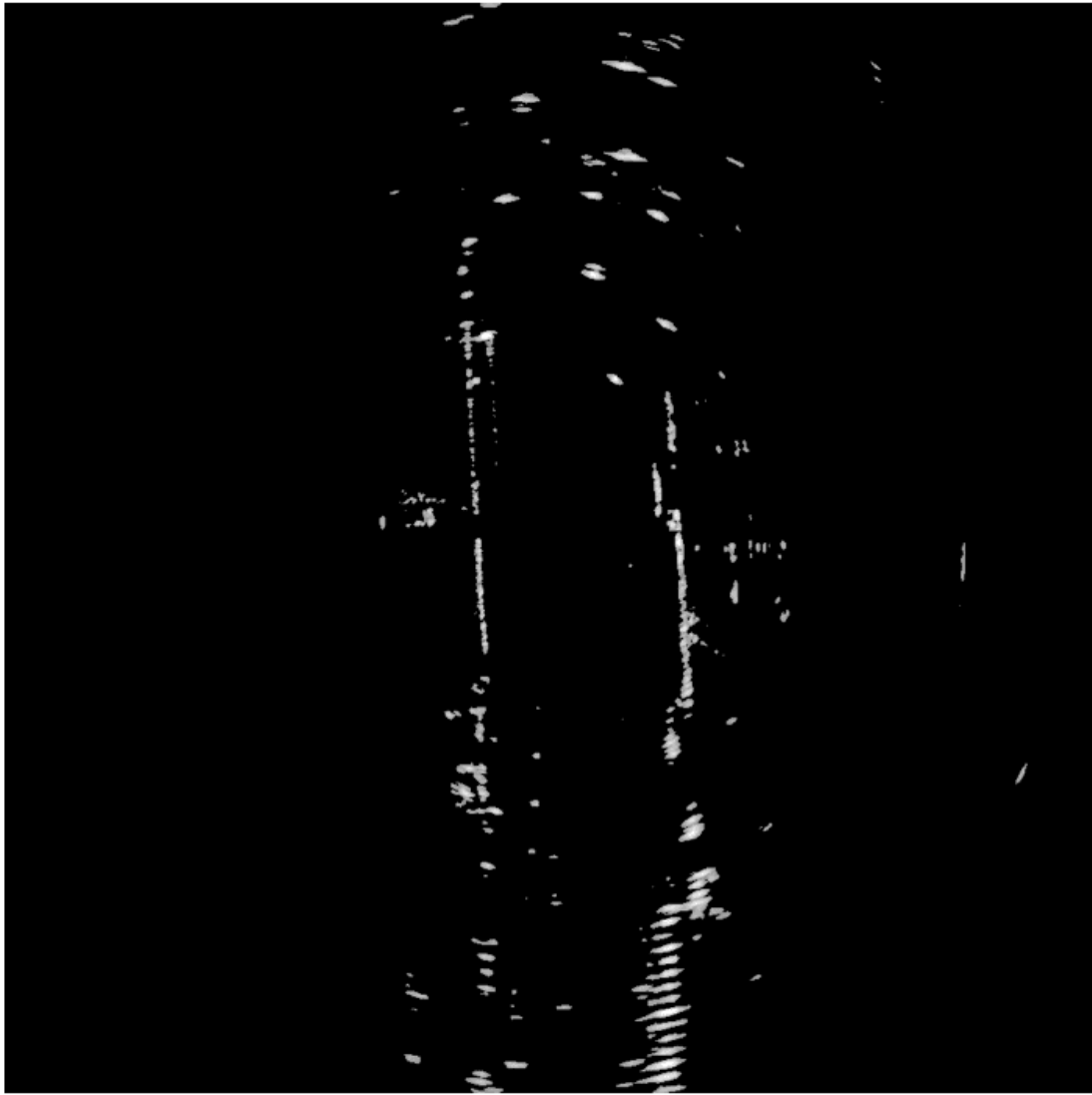}
        \label{fig:dense_nd_map_a}
        }\hspace{-0.14in}
        \subfigure[]{
        \includegraphics[width = 1.1in]{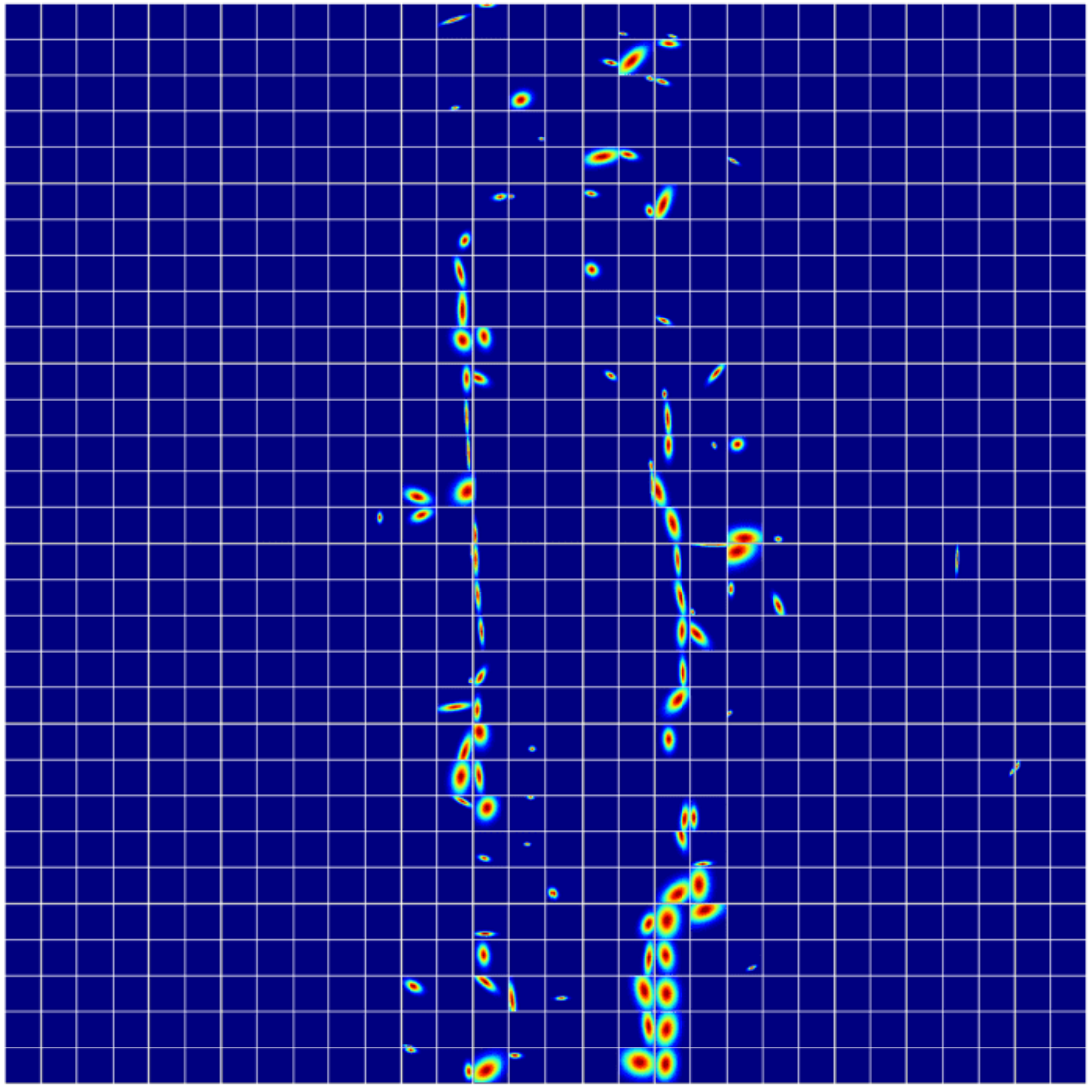}
        \label{fig:dense_nd_map_b}
        }\hspace{-0.14in}
        \subfigure[]{
        \includegraphics[width = 1.1in]{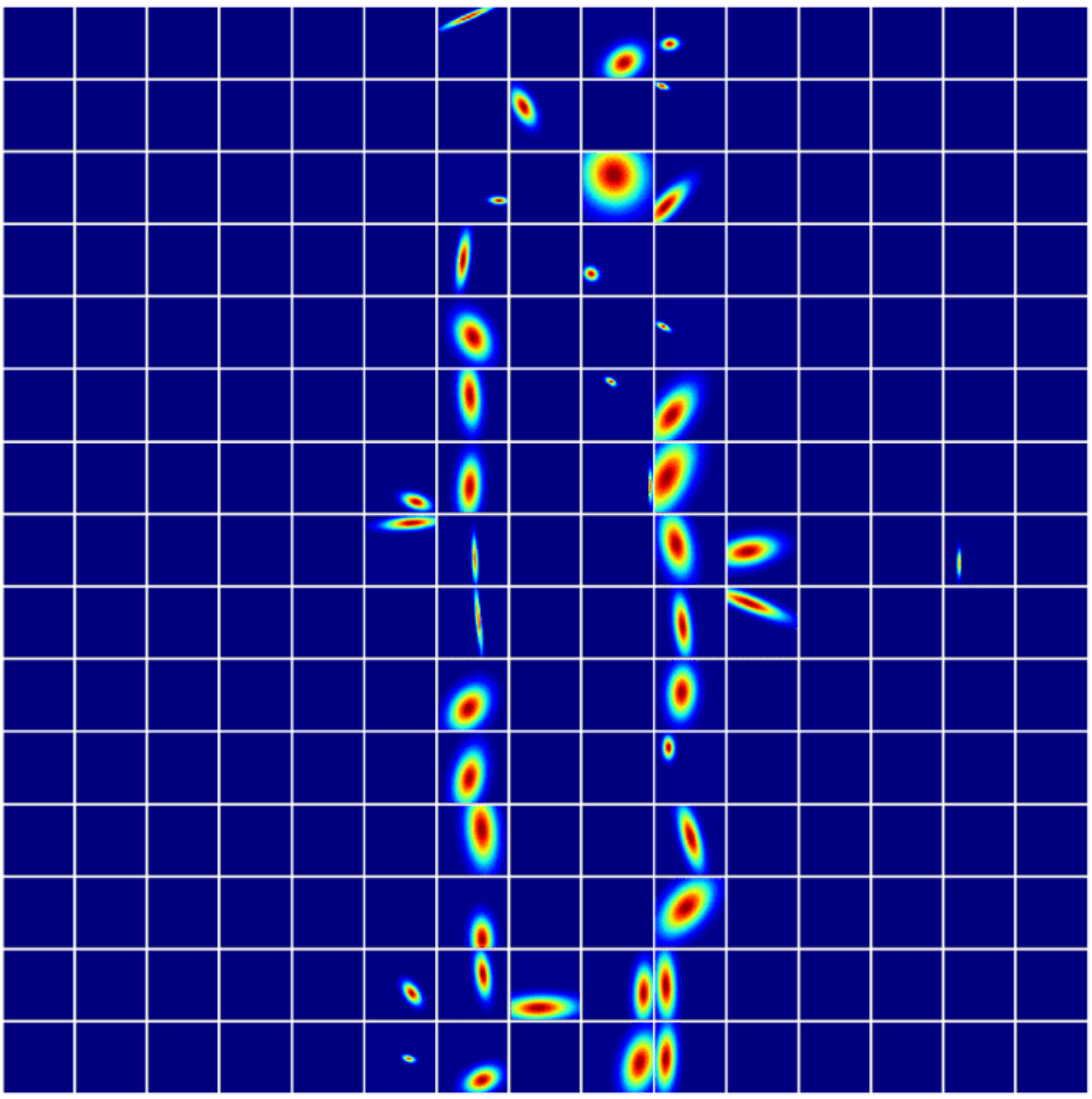}
        \label{fig:dense_nd_map_c}
        }\hspace{-0.14in}
        \caption{Illustration of the normal distribution map created from a scanning radar scan. (a) The thresholded scanning radar scan. (b),(c) The normal distribution map with the grid size of 3.75 m and 7.5 m for visualization.}
        \label{fig:ND Map dense} 
    \end{figure}

\steve{Since automotive radars do not provide returned power of radar points, they are equally weighted ($w_i = 1$, $\forall i$) in automotive RO. Fig.~\ref{fig:ND Map sparse} and Fig.~\ref{fig:ND Map dense} shows the normal distribution maps of an automotive and a scanning radar scan, respectively.}


\subsection{NDT Matching}
\label{sec:Align}

To align two radar scans, the weighted Point-to-Distribution (P2D) NDT matching is applied. 
Inspired by \cite{rapp2015clustering}, which weighted \steve{each radar point in P2D NDT by the similarity of the measured velocity and the estimated velocity at that point,
we design the weight function based on the shifted returned power and uncertainty of each point,
\begin{equation}
 w_i = (p_i - s) e^{-\frac{1}{2}(\sigma_{x}+\sigma_{y})}
\end{equation}
where $\sigma_{x}$ and $\sigma_{y}$ are standard deviation in the covariance matrix corresponding to the radar point. The weight is designed to reduce the weight of those points having high uncertainty. The P2D NDT matching is then achieved by minimizing the cost function using Newton's method \cite{biber2003ndt}.}


\subsection{Discussion}
\label{sec:Discussion}

There are several parameters that need to be set in our algorithm. We analyze how these parameters affect our algorithm.

	\begin{figure}[t]
	\vspace{5pt} 
	\includegraphics[width = 3.4in]{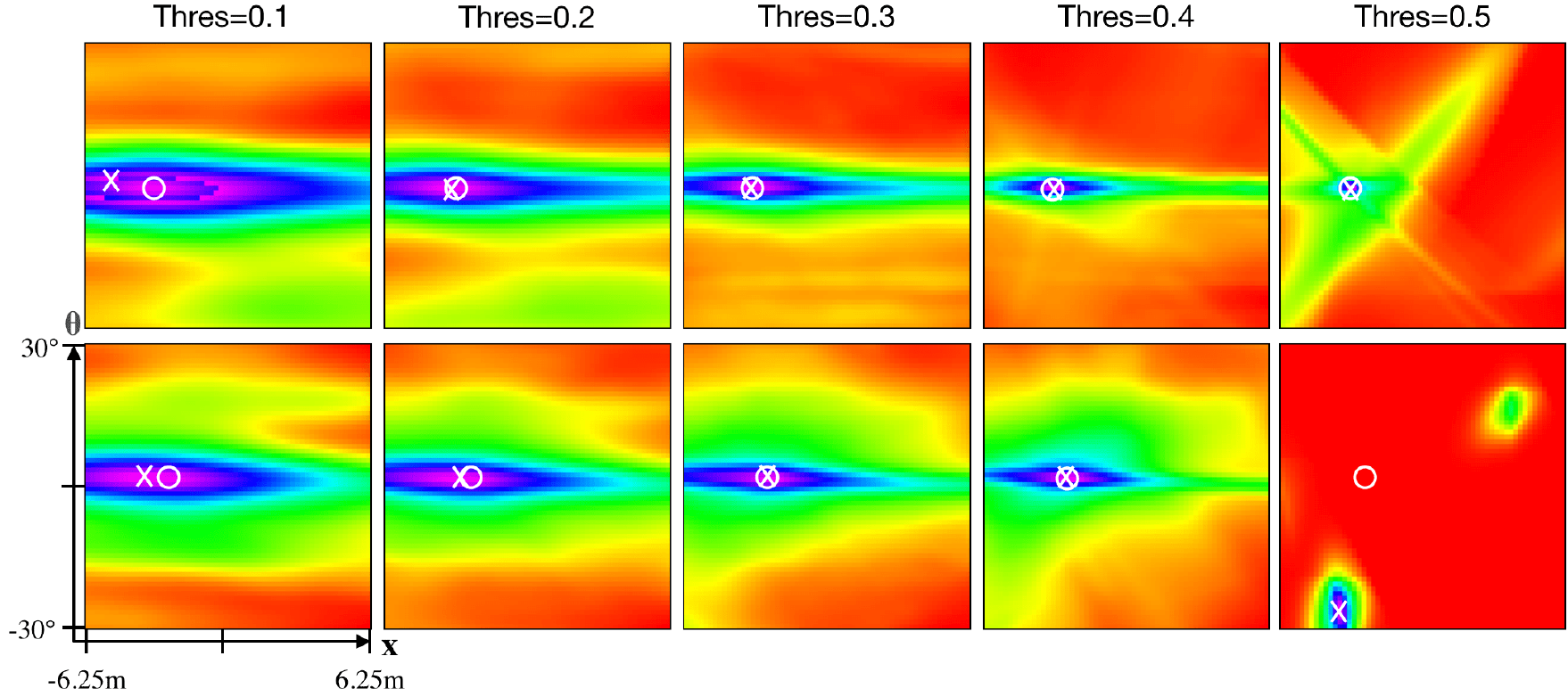} 
	\centering
	\caption{The NDT cost function using different $thres$ value. The circle markers denote the ground truth transformation between two scans, and the cross markers denote the global minimum of each NDT matching.}
	\label{fig:thres_costmap}
	\end{figure}
	\begin{figure}[t]
	\includegraphics[width = 3.4in]{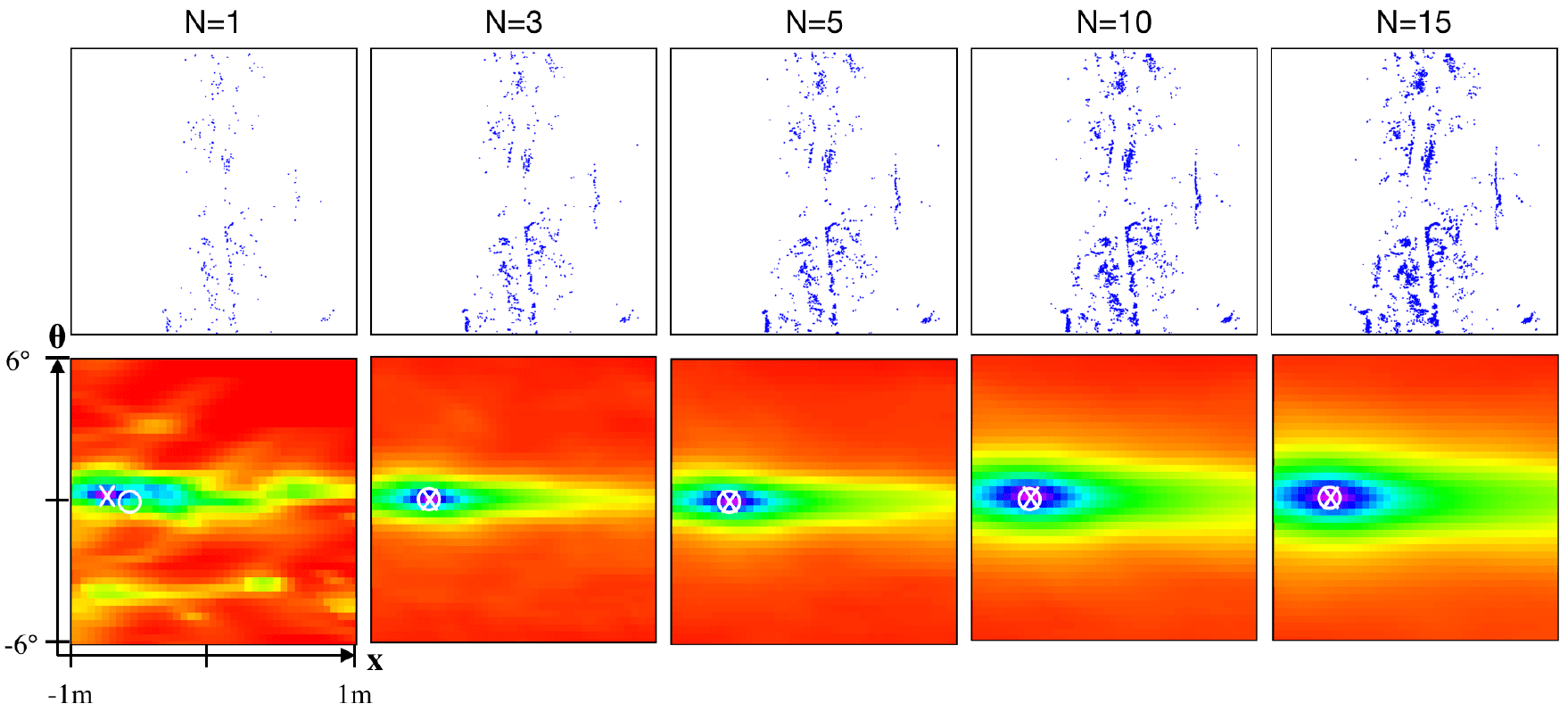} 
	\centering
	\caption{The first row is the radar submap using different $N$ values. The second row is the corresponding NDT cost function. The circle markers denote the ground truth transformation between two scans, and the cross markers denote the global minimum of each NDT matching.}
	\label{fig:stack_costmap}
	\end{figure}

\subsubsection{Threshold value}
\label{sec:discuss thres}
By setting a higher $thres$ parameter, our algorithm removes \steve{more radar points and only consider those} with higher returned power in radar image. This could lead to a more precise outcome due to the noise removal but less robust due to the sparseness of the radar image. 
\steve{NDT cost functions (in $x$ and $\theta$ domains) at different thresholds are shown in Fig.~\ref{fig:thres_costmap}, where $x$ is the vehicle's back and forth movement and $\theta$ is steering angle. At lower thresholds, the cost function has a wider basin of convergence, but its global minimum may not be the correct transformation due to the radar noise effect. 
Note that if the threshold is too high, the global minimum of the cost function could also be incorrect transformation since the radar points would be too sparse or not stationary. In our implementation, we found that a threshold value around 0.3 is proper for maintaining the balance between accuracy and robustness.}

\subsubsection{Number of scans in a Submap}
By setting a higher $N$ parameter, our algorithm uses more radar scans to construct a probabilistic radar submap. This could increase the density of the submap but make our algorithm rely more on ego-velocity. 
The submaps with different $N$'s and their corresponding NDT cost function are illustrated in Fig.~\ref{fig:stack_costmap}. \steve{At $N=1$, the cost function has the narrowest basin of convergence, but its global minimum may not be the correct transformation as the radar scan is too spare to construct a submap.  
At a higher $N$, the cost function has a relatively wider basin of convergence, which implies it is more robust to bad initial guess. However, the accuracy depends more on the accuracy of ego-velocity estimation and ego-velocity covariance estimation.}
	
\subsubsection{Returned Power Shifting}
\begin{figure}[t]
	\vspace{5pt} 
        \centering
        \subfigure[Before power shifting]{
        \includegraphics[width = 1.5in]{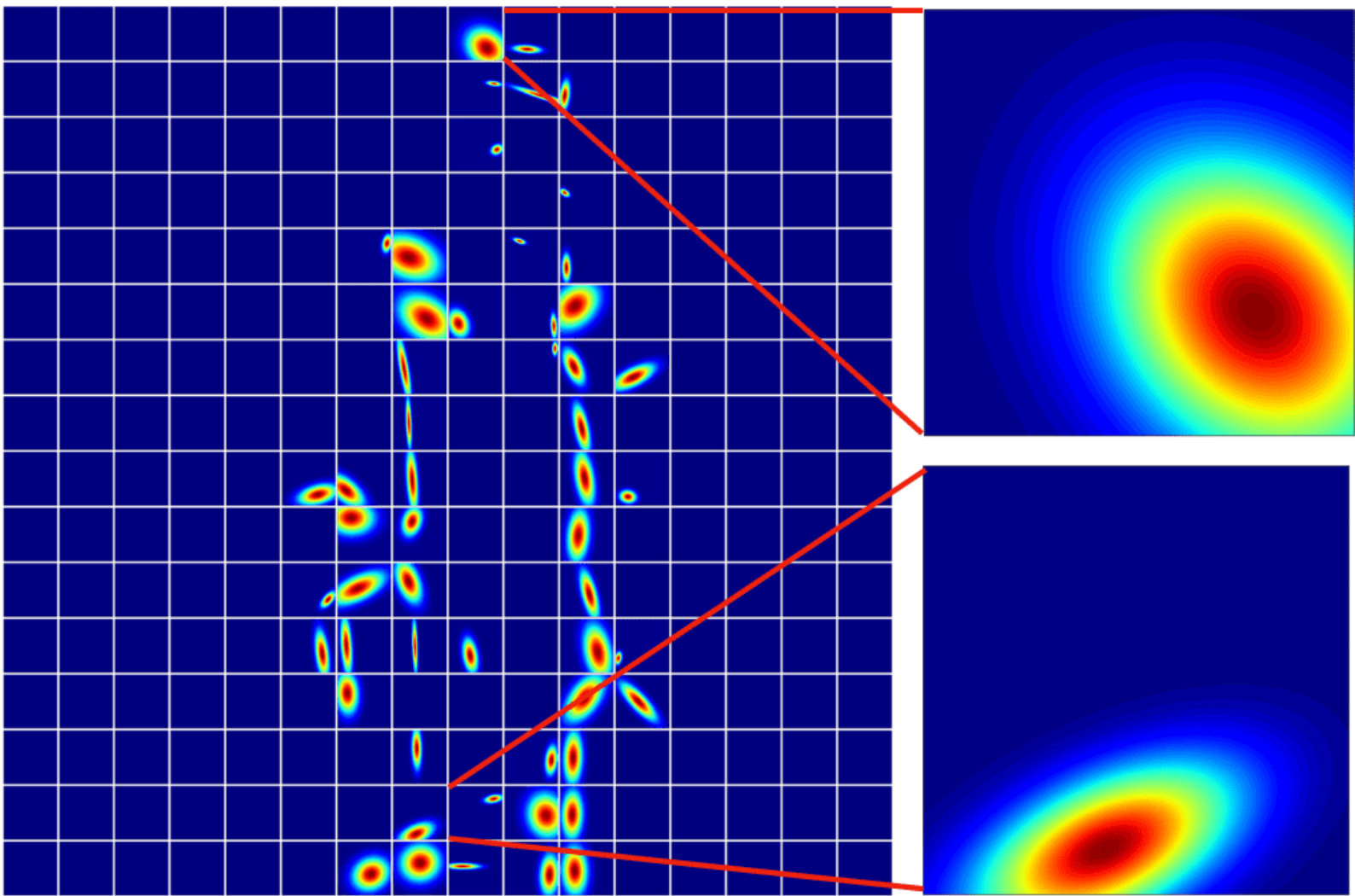}
        }\hspace{-0.14in}
        \subfigure[After power shifting]{
        \includegraphics[width = 1.5in]{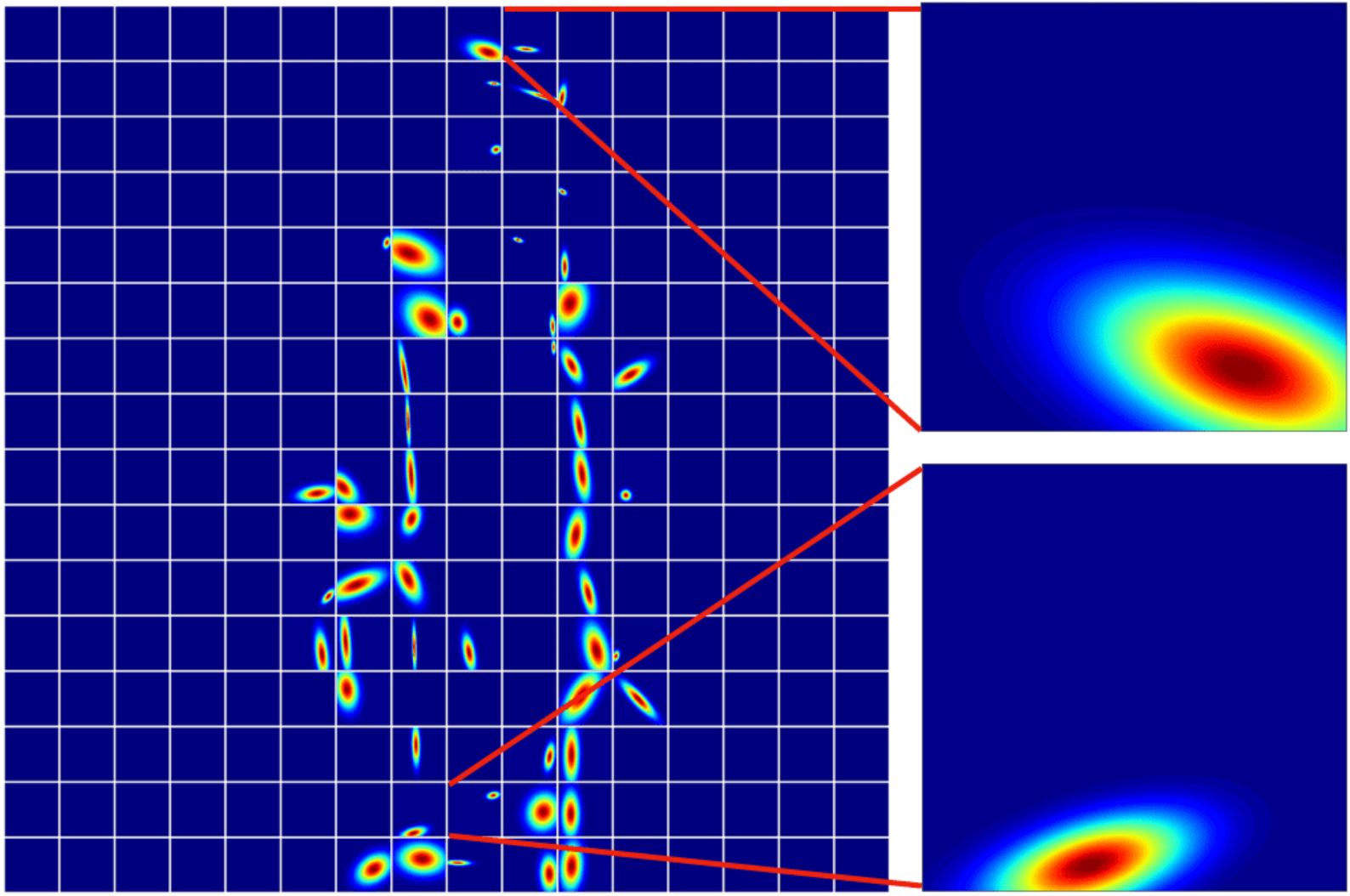}
        }\hspace{-0.0in}
        \caption{The normal distribution map before and after power shifting applied. \steve{The normal distribution of each cell becomes narrower after power shifting.}}
        
        \label{fig:shift_ndmap} 
    \end{figure}

\begin{figure}[t]
        \centering
        \subfigure[Before power shifting]{
        \includegraphics[width = 1.04in]{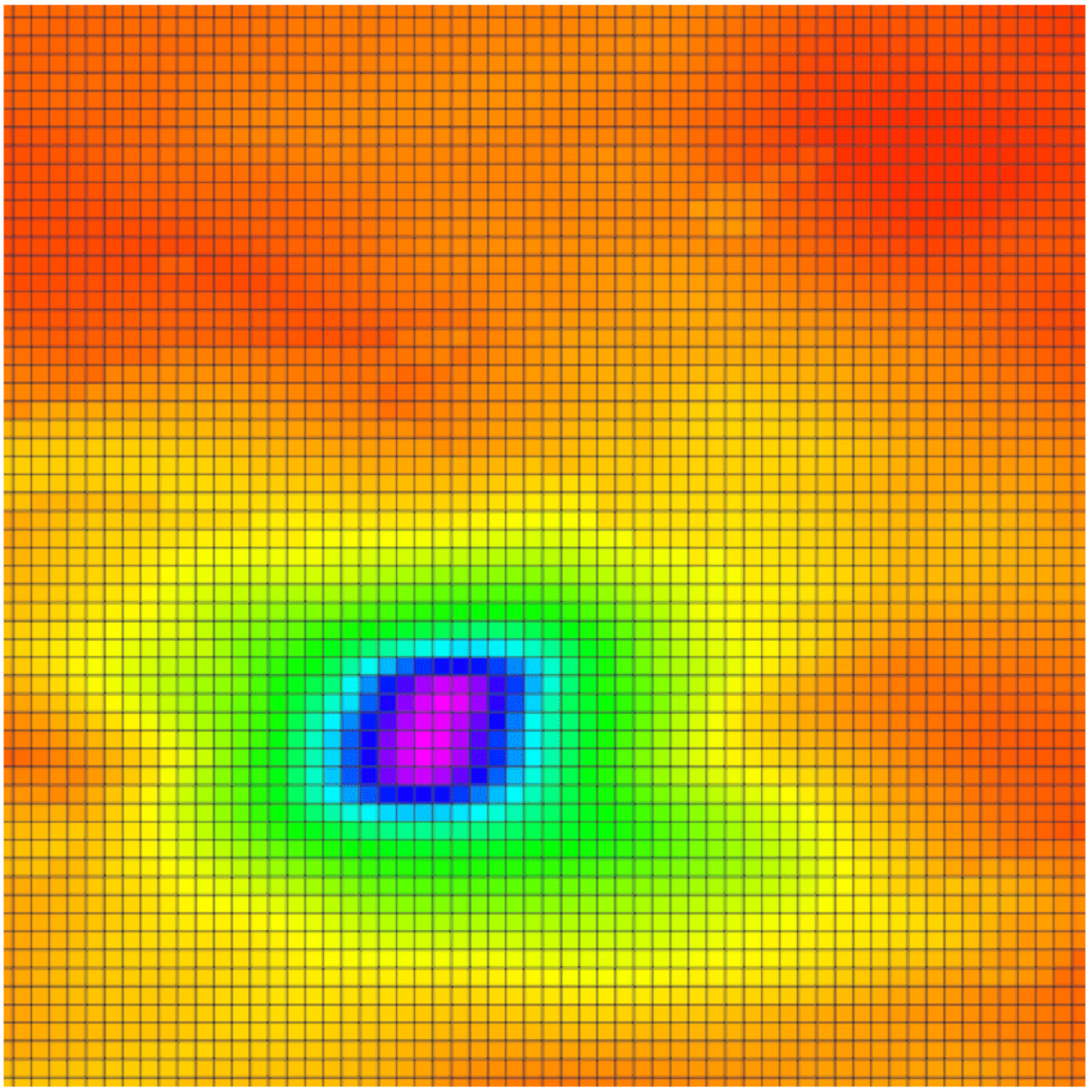}
        }\hspace{0.2in}
        \subfigure[After power shifting]{
        \includegraphics[width = 1.04in]{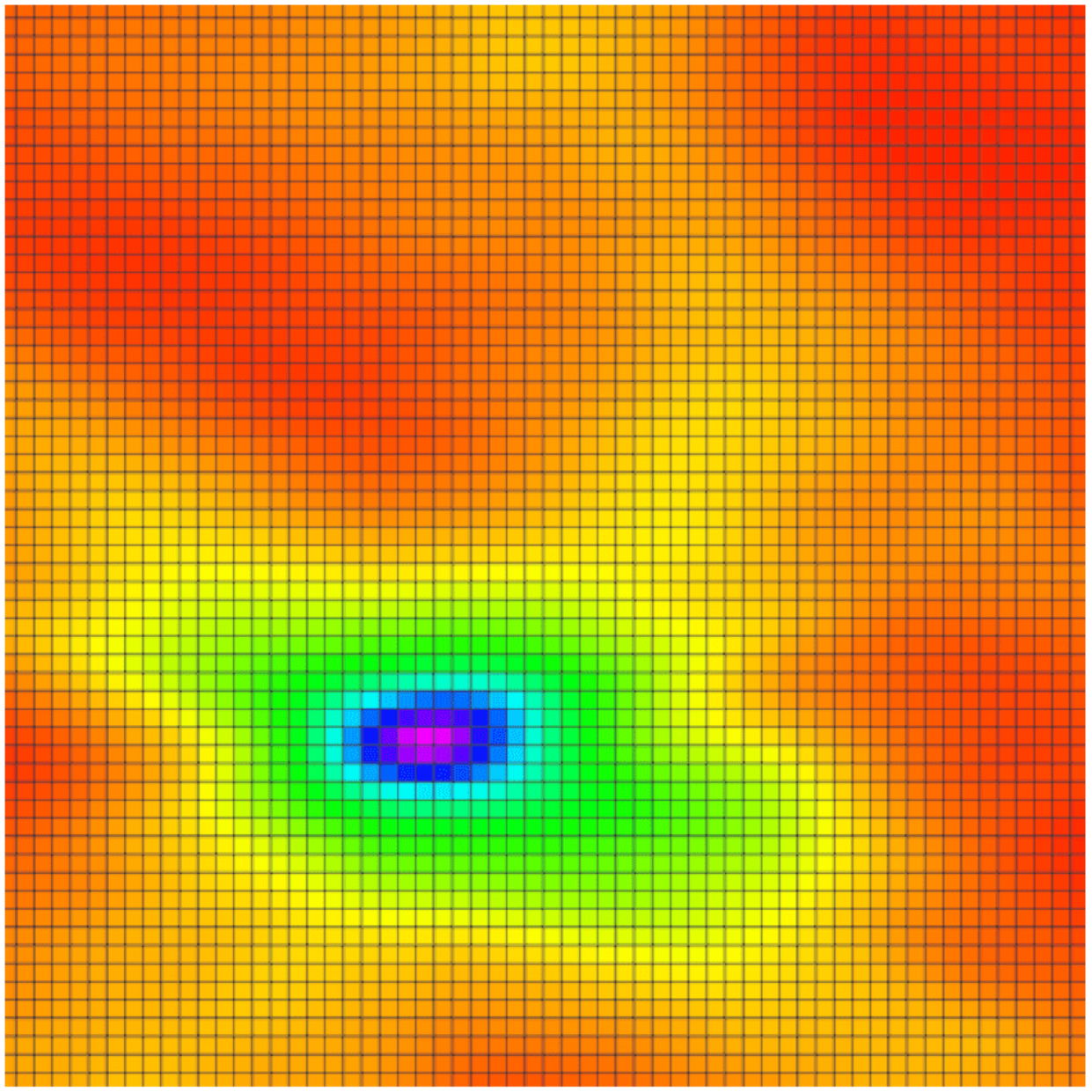}
        }\hspace{0.1in}
        \vspace{-0.2cm}
        \caption{The cost function of NDT before and after power shifting applied. The basin of convergence and region around the global minimum in (b) are narrower than those in (a).}
        \label{fig:shift_costmap} 
    \end{figure}

The parameter $s$ is used to change the weights of radar points based on their returned power. Our algorithm relies more on points with high returned power as $s$ value increases. 
The shifting effect on normal distribution map and the NDT cost function are presented in Fig.~\ref{fig:shift_ndmap} and Fig.~\ref{fig:shift_costmap}, respectively. 
After power shifting, the cost function has a narrower basin of convergence, and NDT can converge to the global minimum more accurately. Table \ref{tab:shift table} shows that the error is reduced by about 40\% after power shifting is applied to the Oxford dataset.


\subsubsection{NDT Grid size}
The grid size of NDT implies the maximal corresponding distance between two scans. At a smaller grid size, NDT could be more robust to moving objects but require a better initial guess. 
The performance of NDT matching is 
\steve{determined by} the size of the grid, but there is no proven way of choosing an optimal grid size. We therefore exploit the motion prior that the system’s maximal acceleration wouldn't exceed $8 m/s^2$ to detect matching failures and increase the grid size when matching fails.

\input{tables/Shift_table}

\section{Experimental Setup \& Odometry evaluation}
\label{sec:datasets}

\subsection{Experimental Setup}

\subsubsection{Automotive Radar Odometry} We test the automotive RO on nuScenes dataset \cite{caesar2020nuscenes}. For radar measurement, we use the radar points within $150m$ and remove outliers (e.g., moving objects or noise) by RANSAC according to the radial velocity measurements. The NDT matching with $g=3.0m$ was adopted in the experiment. 

\subsubsection{Scanning Radar Odometry}
We test the Scanning RO on Oxford Radar RobotCar dataset \cite{barnes2020oxford}. 
For pre-processing, we only use the measurement within $62.5m$ to reduce blurring effect caused by Polar to Cartesian conversion at far distances.
In the experiment, we set the parameters $g=3.75m$, $thres=0.333$, and $shift=0.333$ at $0.125m/pixel$ resolution.
\steve{The grid size is increased by $2.5m$ whenever the matching failure occurs. If the matching still fails when $g$ is greater than $12.5m$, then the failure might be caused by moving objects. In this case, the $s$ value is halved to reduce the influence of moving objects, particularly buses and trucks with high returned power.} 
Moreover, to reduce the influence of moving objects when the car is stopping, we set $g=1.5m$ when the estimated ego-velocity is lower than $1.5km/hr$.

\subsection{Odometry Evaluation}
For the scanning RO evaluation, we follow the KITTI odometry benchmark \cite{geiger2012we}, which computes the mean translational and rotational errors from length $100m$ to $800m$ with a $100m$ increment for each radar pose. The final error is the average of all these errors; For the automotive RO evaluation, we only compute the mean error in $1m$ because the ground truth of pose in the nuScenes dataset is discontinuous in long distance.

\section{Results}
\label{sec:results}

\subsection{Automotive Radar Odometry}

\begin{figure}[!t]
	\vspace{2pt} 
	\includegraphics[width = 3.4in]{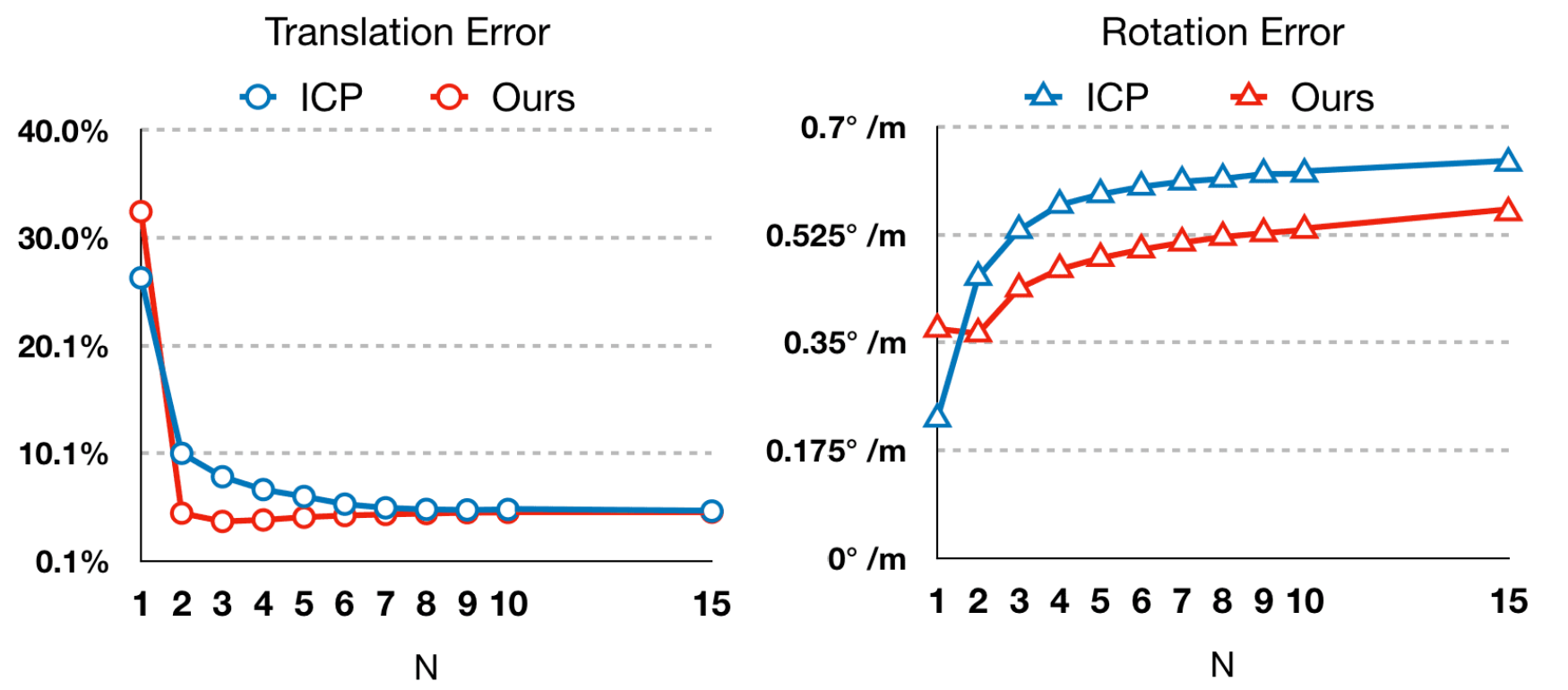}
	\centering
	\caption{The figure shows the translational and rotational error of ICP and our approach using different $N$ values.}
	\label{fig:N_result_plot}
	\end{figure}

The results of an automotive RO is shown in Table \ref{tab:sparse ro result table} with 13Hz odometry frequency. We compare our approach with convention ICP used in \cite{ward2016vehicle}, submap ICP \cite{holder2019real}, Kellner et al.'s ego-velocity integration \cite{kellner2014instantaneous}, and Rapp et al.'s joint-Doppler NDT-based method \cite{rapp2017probabilistic}. 
For a fair comparison, we implemented grid-based NDT (instead of clustering-based NDT) in both Rapp et al.'s method \cite{rapp2017probabilistic} and our method. 
In the table, we also show the results obtained with different $N$'s. \steve{Besides, the error plots of ICP and our method at different $N$'s} are shown in Fig.~\ref{fig:N_result_plot}.
At $N=1$, the conventional single-scan matching has relatively better rotation estimation but worse translation estimation. The main reason is a single radar scan is too sparse, 
so the accuracy of scan matching \steve{is simply determined by} the sensor's accuracy, 
which is $0.1m$-$0.4m$ in translation and $0.1^{\circ}$-$1^{\circ}$ in rotation.
Compared with single scan matching, the submap matching provides better translation estimation but worse rotation estimation. This is because even the submap matching improves the translation estimation by constructing denser information, it is still affected by the 
\steve{accuracy of ego-velocity estimation}, which has a relatively higher rotation error. Although the results of both ICP and our approach have better-estimated translation and worse-estimated rotation as $N$ increases, the proposed method always outperforms ICP in submap matching because our method considers uncertainties of measurements and ego-velocity. 
Our approach has the best translation result at $N=3$. \steve{This can be explained as follows. Because the quality of the probabilistic submap depends on ego-velocity covariance estimation and the covariance estimation error accumulates as $N$ increases, the translational error of our results }
becomes closer to the translational error of ICP if the $N$ value is too high. For instance, the translation error of our method is almost the same as ICP at $N=15$.



\subsection{Scanning Radar Odometry}

Table \ref{tab:dense ro result table} shows the results of our RO and three other RO's \steve{tested on the Oxford Radar RobotCar Dataset, including the ``Masking by Moving'' \cite{barnes2019masking}, ``RO Cen'' \cite{ro_cen2}, and ``RadarSLAM'' \cite{hong2020radarslam}. 
}
The resolution and error statistics are excerpted from \cite{barnes2019masking} and \cite{hong2020radarslam}. For fair comparison, we choose spatial cross-validation results of ``Masking by Moving'' shown in appendix of \cite{barnes2019masking} which evaluate in unseen scenes instead of the scenes same as training data.
Overall, our approach has the smallest error and outperforms the state-of-the-art, ``Masking by Moving'', by reducing translational and rotational error by 29\% and 27\%, respectively. 
Although our approach utilized simple thresholding for pre-processing instead of a trained masking network, the proposed method has better outlier handling by using grid-based NDT compared to correlation-based scan matching used in \cite{barnes2019masking}. The correlation-based scan matching in \cite{barnes2019masking} could be affected by outliers if the masking network did not remove all outliers. However, the proposed NDT method could handle the outliers which the thresholding module did not remove. 
While the result of \cite{barnes2019masking} is tested at a relatively low resolution, we do not know the outcome of \cite{barnes2019masking} using an equivalent resolution setting.

\input{tables/Sparse_RO_Result_Table}

\input{tables/Dense_RO_Result_Table}

\subsection{Compare Automotive RO with Scanning RO}
\label{sec:automotive_vs_scanning}
\steve{We compare the accuracy of automotive RO and scanning RO on the  frame-by-frame errors (the bottom of Table \ref{tab:Experiment}) because the per-meter errors could be affected by the update frequency of sensors.} 
\steve{The frame-by-frame errors in Table \ref{tab:Experiment} show that our automotive RO could achieve comparable odometry accuracy as scanning RO despite that the automotive radar has relative lower accuracy and resolution.}
\steve{We consider a possible reason} is that although the scanning radar provides radar images with relatively higher resolution, the radar images are blurry, which only provides rough positions of objects. The blurry measurements represents the uncertainty of measurement, just like the uncertainty of radar points measured by automotive radar.
However, we cannot precisely compare automotive and scanning RO \steve{because the two datasets have different scenarios and ground truth of reference odometry.} The ground truth provided by nuScenes is the lidar localization result using the lidar HD map. In contrast, the GPS/INS solution has been used as the ground truth of the Oxford dataset in this experiment because we found that the SE2 RO ground truth in the Oxford dataset is not sufficiently robust compared to GPS/INS solution.

\subsection{Compared with Lidar Odometry}
\label{sec:radar_vs_Lidar}
Results at the bottom of Table \ref{tab:Experiment} shows the radar and lidar odometry results on two datasets. To compare the radar and lidar odometry, we downsample all sensors to approximately 4Hz and evaluate the odometry performance by the frame-by-frame error.
The lidar odometry is achieved by the fine-tuned multi-stage ICP, and lidar points on the ego-vehicle and ground are removed by a fixed spatial threshold. To have an fair comparison with radar, we only calculate the SE2 error of lidar odometry. The results show that both the automotive and scanning radar achieves centimeter-level accuracy like lidar odometry. 


\section{Conclusion and Future Work}
\label{sec:conclusion}
In this paper, we present state-of-the-art RO, which is designed for both scanning and automotive radars. The proposed method is easy yet efficient. By introducing the straightforward thresholding and probabilistic radar submap, our method can both remove radar noise from scanning radar scans and construct dense information from automotive radar scans. 
The RO is achieved by consecutive scan matching using the proposed weighted probabilistic NDT, which takes the returned power and the uncertainties of radar points into account.
\steve{The experimental results on two public datasets} demonstrate that our RO achieves centimeter-level accuracy as lidar odometry does, and automotive RO \steve{could reach a similar level of accuracy as} scanning RO. To the best of our knowledge, our work is the first RO method that can operate on both types of radars, and the first one makes a odometry comparison between them and with lidar. 

In the future, we aim to evaluate automotive and scanning RO on \steve{the same} dataset for a \steve{fair} comparison. Also, we intend to extend our approach to 4D radars to achieve 6 degrees of freedom RO and investigate the fusion of IMU to improve RO.



\bibliographystyle{IEEEtran}
\bibliography{references}

\end{document}

%% file: tables/Radar_Table.tex
\setlength{\arrayrulewidth}{0.2mm}
\setlength{\tabcolsep}{1.7pt}
\renewcommand{\arraystretch}{1.6}

\begin{table}[!t]
\begin{center}
\scalebox{0.9}{
\begin{tabular}{|c|c|c|}
\hline
                    & Automotive radar      & Scanning radar \\ \hline
                    & Continental ARS 408-21         & Navtech  CTS350-X       \\ \hline
Range resolution    & 0.39-1.79 m                      & 0.0438 m*                 \\ \hline
Range accuracy      & 0.1-0.4 m                  & --                      \\ \hline
Azimuth resolution & $1.6^{\circ}$-$4.5^{\circ}$ & \begin{tabular}[c]{@{}c@{}}scanning resolution: $0.9^{\circ}$*\\ azimuth beamwidth: $2^{\circ}$*\end{tabular} \\ \hline
Azimuth accuracy    & $0.1^{\circ}$-$1^{\circ}$ & --                      \\ \hline
Frequency           & 13 Hz                          & 4 Hz*                    \\ \hline
Maximum range       & 250 m                          & 163 m*                   \\ \hline
Doppler information & $\pm0.1$km/h                   & No                      \\ \hline
\end{tabular}
}
\end{center}
{\raggedright 
*These value can be varied according to the different configuration.
\par}
\vspace{-1mm}
\caption{Automotive and Scanning radar.}
\label{tab:radar table}
\vspace{-1mm}
\end{table}

%% file: tables/Experiment.tex
\setlength{\arrayrulewidth}{0.2mm}
\setlength{\tabcolsep}{1.7pt}
\renewcommand{\arraystretch}{1.6}
\begin{table}[t]
\centering
\scalebox{0.9}{
\addvbuffer[8pt]{ 
\begin{tabular}{clccccccc}
\hline
Dataset                           &                      & \multicolumn{3}{c}{nuScenes}                                           &                      & \multicolumn{3}{c}{Oxford}                             \\ \cline{1-1} \cline{3-5} \cline{7-9} 
Type of radar                     &                      & \multicolumn{3}{c}{Automotive Radar}                                   & \multicolumn{1}{l}{} & \multicolumn{3}{c}{Scanning Radar}                     \\ \cline{1-1} \cline{3-5} \cline{7-9} 
\multirow{2}{*}{RO Benchmarks}    &                      & Trans. Error                  &                      & Rot. Error      &                      & Tran. Error   &                      & Rot. Error      \\
                                  &                      & (\%)                          & \multicolumn{1}{l}{} & (deg/m)         & \multicolumn{1}{l}{} & (\%)          & \multicolumn{1}{l}{} & (deg/m)         \\ \cline{1-1} \cline{3-3} \cline{5-5} \cline{7-7} \cline{9-9} 
SOTA Automotive RO \cite{holder2019real}                &                      & 7.91                          &                      & 0.5325          &                      & --            &                      & --              \\
SOTA Scanning RO \cite{barnes2019masking}                  &                      & --                            &                      & --              &                      & 2.78 &                      & 0.0085          \\
Proposed RO                &                      & \textbf{3.85}                 &                      & \textbf{0.4430} &                      & \textbf{1.96}          &                      & \textbf{0.0060} \\ \hline
\multirow{2}{*}{Lidar Benchmarks} &                      & Trans. Error                  &                      & Rot. Error      &                      & Tran. Error   &                      & Rot. Error      \\
                                  & \multicolumn{1}{c}{} & \multicolumn{1}{l}{(m/frame)} &                      & (deg/frame)     &                      & (m/frame)     &                      & (deg/frame)     \\ \cline{1-1} \cline{3-3} \cline{5-5} \cline{7-7} \cline{9-9} 
Lidar ICP                         &                      & 0.0240                        &                      & 0.0488          &                      & 0.0617        &                      & 0.0448          \\
Proposed RO                &                      & 0.0355                        &                      & 0.1043          &                      & 0.0652        &                      & 0.0736          \\ \hline
\end{tabular}
}
}
\caption{\label{tab:Experiment} Performance of Proposed Radar Odometry.}
\vspace{-0mm}
\end{table}

%% file: tables/Shift_table.tex
\setlength{\arrayrulewidth}{0.2mm}
\setlength{\tabcolsep}{3pt}
\renewcommand{\arraystretch}{1.5}

\begin{table}[t]
\begin{center}
\scalebox{0.85}{
\addvbuffer[8pt]{ 
\begin{tabular}{clclc}
\hline
Our approach                         &  & Trans. Error        &  & Rot. Error          \\
\multicolumn{1}{l}{}                 &  & (\%)                     &  & (deg/m)                 \\ \cline{1-1} \cline{3-3} \cline{5-5}
Before power shifting      &  & 4.2232                   &  & 0.0131                  \\
After power shifting         &  & \textbf{2.4859}                   &  & \textbf{0.0073}                  \\ \hline
\end{tabular}
}
}
\end{center}
\caption{\label{tab:shift table} Power shifting effect on scanning radar odometry.}
\vspace{-3mm}
\end{table}

%% file: tables/Sparse_RO_Result_Table.tex
\setlength{\arrayrulewidth}{0.2mm}
\setlength{\tabcolsep}{1.5pt}
\renewcommand{\arraystretch}{1.6}

\begin{table}[t]
\begin{center}
\scalebox{0.9}{
\addvbuffer[6pt]{ 
\begin{tabular}{cccccccccclccclc}
\hline
 $N$              & \multicolumn{3}{c}{ICP \cite{ward2016vehicle,holder2019real}}          &  & \multicolumn{3}{c}{Our Approach}     &  & \multicolumn{3}{c}{Kellner et al.'s \cite{kellner2014instantaneous}} &  & \multicolumn{3}{c}{Rapp et al.'s \cite{rapp2017probabilistic}} \\ \cline{2-4} \cline{6-8} \cline{10-12} \cline{14-16} 
               & Trans.  &  & Rot.        &  & Trans.     &  & Rot.        &  & Trans.      &       & Rot.     &  & Trans.    &     & Rot.   \\
               & (\%)        &  & (deg/m)         &  & (\%)            &  & (deg/m)         &  & (\%)             &       & (deg/m)      &  & (\%)           &     & (deg/m)    \\ \cline{2-2} \cline{4-4} \cline{6-6} \cline{8-8} \cline{10-10} \cline{12-12} \cline{14-14} \cline{16-16} 
 1   & 26.3340     &  & \textit{0.2264} &  & 32.4595         &  & 0.3728          &  & 6.3370           &       & 0.8125       &  & 24.3089        &     & 0.3826     \\
 3   & 7.9140      &  & 0.5325          &  & \textbf{3.8455} &  & \textbf{0.4430} &  &                  &       &              &  &                &     &            \\
 5   & 6.0893      &  & 0.5907          &  & 4.1868          &  & 0.4843          &  &                  &       &              &  &                &     &            \\
 10  & 4.8887      &  & 0.6241          &  & 4.5887          &  & 0.5305          &  &                  &       &              &  &                &     &            \\ \hline
\end{tabular}
}
}
\caption{\label{tab:sparse ro result table} Radar odometry on NuScenes Dataset.}
\end{center}
\vspace{-2mm}
\end{table}

%% file: tables/Dense_RO_Result_Table.tex
\setlength{\arrayrulewidth}{0.2mm}
\setlength{\tabcolsep}{1.6pt}
\renewcommand{\arraystretch}{1.6}

\begin{table}[t]
\begin{center}
\scalebox{0.9}{
\begin{tabular}{clclclc}
\hline
Benchmarks                           &  & Resolution &  & Trans. Error        &  & Rot. Error          \\
\multicolumn{1}{l}{}                 &  & (m/pixel)  &  & (\%)                     &  & (deg/m)                 \\ \cline{1-1} \cline{3-3} \cline{5-5} \cline{7-7} 
RO in RadarSLAM \cite{hong2020radarslam}       &  &  0.0432     &  &  3.0173                     &  & 0.0091                    \\
RO Cen Full Res \cite{ro_cen2}       &  & 0.0432     &  & 6.3813                   &  & 0.0189                  \\
RO Cen Equiv. \cite{ro_cen2}         &  & 0.1752     &  & 3.6349                   &  & 0.0096                  \\
Masking by Moving \cite{barnes2019masking}  &  & 0.4        &  & 2.7848                   &  & 0.0085         \\
Our approach                       &  & 0.125      &  & \textbf{1.9584}          &  & \textbf{0.0060}                  \\\hline
\end{tabular}
}
\end{center}
\caption{\label{tab:dense ro result table} Radar odometry on Oxford Radar RobotCar Dataset.}
\vspace{-2mm}
\end{table}